\documentclass[
	language=english, 
	type=master, 
]{isthesis}

\usepackage{tikz}

\usetikzlibrary{calc}
\usetikzlibrary{matrix}
\usetikzlibrary{positioning}
\usetikzlibrary{shapes.geometric}

\usepackage{booktabs}
\usepackage{multirow}
\usepackage{lscape}

\newacronym[shortplural={KMUen}, longplural={Kleine und Mittlere Unternehmen}]{kmu}{KMU}{Kleines und Mittleres Unternehmen}
\newacronym{CD}{CD}{Corporate Design}
\newacronym{SQL}{SQL}{Structured Query Language}
\newacronym{FAU}{FAU}{Friedrich-Alexander-Universit\"at Erlangen-N\"urnberg}
\newacronym{BPM}{BPM}{Business Process Management}
\newacronym{npm}{NPM}{Node Package Manager}
\newacronym{diss}{DISS}{Digital Industrial Service System}

\addsymboltolist{$\Pi$}{Pi}{Projection}
\addsymboltolist{$\Join$}{Join}{Natural Join}
\addsymboltolist{$\sigma$}{Selection}{Selection}

\isthesis{
    title={Efficient Compression of Multitask Multilingual Speech Models},
    author-name={Thomas PALMEIRA FERRAZ}, 
    author-email={thomas.palmeira@telecom-paris.fr},
    author-city={Paris, France},
    principal-supervisor={Prof. Dr. Emmanuel DUPOUX (PSL/Meta AI)}, 
    tutor-supervisor={Dr. Marcely ZANON BOITO (NAVER LABS Europe) 
    \\
    $\qquad \qquad \qquad \qquad \qquad \quad$ \: \,\,Dr. Caroline BRUN (NAVER LABS Europe)
    \\
    $\qquad \qquad \qquad \qquad \qquad \quad$ \: \,\,Dr. Vassilina NIKOULINA (NAVER LABS Europe)},
    group={Master of Science - Mathématiques, Vision, Apprentissage (MVA)},
    seminar={Scientific Writing for Beginners}, 
    submission-date={2023-10-02} 
}

\begin{document}
    \newcounter{savepage}
    \maketitle

    \begin{abstract}

     \textit{Whisper} is a multitask and multilingual speech model covering 99 languages. It yields commendable automatic speech recognition~(ASR) results in a subset of its covered languages, but the model still underperforms on a non-negligible number of under-represented languages, a problem exacerbated in smaller model versions. In this work, we examine its limitations, demonstrating the presence of speaker-related (gender, age) and model-related (resourcefulness and model size) bias. Despite that, we show that only model-related bias are amplified by quantization, impacting more low-resource languages and smaller models. Searching for a better compression approach, we propose \textit{DistilWhisper}, an approach that is able to bridge the performance gap in ASR for these languages while retaining the advantages of multitask and multilingual capabilities. Our approach involves two key strategies: lightweight modular ASR fine-tuning of \texttt{whisper-small} using language-specific experts, and knowledge distillation from \texttt{whisper-large-v2}. This dual approach allows us to effectively boost ASR performance while keeping the robustness inherited from the multitask and multilingual pre-training. Results demonstrate that our approach is more effective than standard fine-tuning or LoRA adapters, boosting performance in the targeted languages for both in- and out-of-domain test sets, while introducing only a negligible parameter overhead at inference.
	\end{abstract}
    
    \tableofcontents


    


	
	%

    \setcounter{savepage}{\value{page}}

    \begin{content}
		\chapter{Introduction}

\section{Motivation}

Over the past three years, the field of Natural Language Processing (NLP) has been revolutionized by the introduction of large pre-trained models, often referred to as "foundation models." These models, both for text and speech, are trained on vast amounts of unlabeled data and can subsequently be fine-tuned for specific tasks using limited labeled data.

Multilingual foundation models have garnered significant attention due to their ability to handle hundreds of languages within a single model. However, they face a challenge known as the \textit{curse of multilinguality}: in order to maintain high performance across all supported languages, these models require an increase in the number of parameters, leading to larger memory requirements and slower inference times. This can render the use of such models impractical in certain scenarios. To address this issue, research has been conducted on model compression techniques, although these methods may inadvertently exacerbate biases present in the model.

This internship focuses on OpenAI's Whisper, a family of multilingual multi-task speech models known for their impressive performance in speech recognition. These models exhibit robustness when transcribing speech recorded under various conditions, surpassing the capabilities of previous models.

However, there remain important questions to explore regarding Whisper and its multi-task learning approach. Although the model presents exceptional capability for transcribing and translating English, its performance in other languages indicates a decline in multilingual capabilities as the model size decreases. Additionally, we aim to investigate how this multilingual architecture handles biases related to different speakers, including gender, age, and accent. These questions drive our research to enhance the understanding of Whisper's capabilities and limitations.

\section{Internship Objectives}

This internship has three main objectives:
\begin{enumerate}
    \item Conduct a comprehensive analysis of bias within the Whisper model family, with a specific focus speaker-related (gender, age, accent) and model-related (model size, resourcefulness, similar languages) biases;
    \item Explore how light compression techniques, such as quantization, may either mitigate or exacerbate any identified biases within the Whisper models; 
    \item Propose a better compression approach that effectively reduces any disparities found in the models.
\end{enumerate}

\section{Contributions of this work}

This work offers two significant contributions. Firstly, it provides a comprehensive analysis of the biases present in the Whisper model and examines how quantization impacts these biases. Secondly, it introduces an alternative model compression method called \textit{DistilWhisper}, which enhances the performance of smaller Whisper models. Additionally, all models and code developed in this research will be made available as open-source resources.

The structure of this report is as follows: Chapter \ref{chap:background} provides essential fundamentals and a comparison with related work to establish a foundational understanding. Chapter \ref{chap:bias} details the experimental setup and results of the investigation into bias when quantizing Whisper. This investigation leads to the proposal of \textit{DistilWhisper}, in Chapter \ref{chap:distilwhisper}, a novel parameter-efficient distillation approach that leverages small pre-trained models. Chapter \ref{chap:results} covers the validation of the proposed approach, as well as some interesting analysis. Finally, Chapter \ref{chap:conclusion} summarizes the primary findings and conclusions of this work.



\section{About NAVER LABS Europe}

NAVER LABS is the R\&D subsidiary of NAVER, Korea’s leading internet company and the part of NAVER responsible for creating future technology. Its world-class researchers in Korea and Europe create new connections between people, machines, spaces and information by advancing technology in AI, robotics, autonomous driving, 3D/HD mapping and AR.

NAVER LABS Europe is the biggest industrial research lab in artificial intelligence in France and a hub of NAVER’s global AI R\&D Belt, a network of centers of excellence in Korea, Japan, Vietnam, USA \& Europe. The scientists at NAVER LABS Europe conduct fundamental and applied research in machine learning (optimization, robotics), computer vision, natural language processing and UX and ethnography. The site is located in Grenoble, France.
		\chapter{Background and Related Work} \label{chap:background}

\section{State of the Art for Automatic Speech Recognition} \label{sec:sota-asr}

Current ASR approaches primarily involve adapting pre-trained Transformer stacks \citep{vaswani2017attention}, which are initially trained through self-supervised learning (SSL) on unlabeled audio data. These pre-trained models can vary in their use of pre-text tasks (e.g., wav2vec 2.0~\citep{baevski2020wav2vec}, HuBERT~\citep{hsu2021hubert}, WavLM \citep{chen2022wavlm}) and the range of languages they cover (e.g., XLSR-53~\citep{conneau21_interspeech}, XLS-R~\citep{babu22_interspeech}, MMS~\citep{pratap2023scaling}, Google-USM~\citep{zhang2023google}). This development of models has also seen the introduction of monolingual and multilingual SSL benchmarks. Examples of such benchmarks include SUPERB for English~\citep{yang21c_interspeech}, LeBenchmark~\citep{evain2021task} for French, and ML-SUPERB ~\citep{shi2023ml}, which covers 143 languages. 

In contrast to this line of research, the Whisper model relies on weak supervision, meaning it is trained solely on weakly labeled data (without self-supervision). Nevertheless, with an ample amount of data, the Whisper model achieves competitive results when compared to monolingual~\citep{radford2023robust,gandhi2022esb} and multilingual~\citep{pratap2023scaling} SSL models. More details about Whisper can be found on Section~\ref{sec:Whisper}. For broader ASR benchmarks, facilitating comparisons between SSL pre-training and multitasking weakly-supervised training, the ESB benchmark from HuggingFace~\citep{gandhi2022esb} for English is an illustrative example.

\section{Domain Adaptation}

Domain adaptation consist in the process of adapting a pre-existing trained model to a new domain or task with minor weight adjustments, rather than retraining the entire model from scratch. In the past, this adaptation was primarily carried out through full fine-tuning, where all the model's weights were updated. In the case of Transformer-based models, it is also common to proceed adaptation choosing to update only specific layers, usually the final ones~\citep{10.1162/coli_a_00434}.

More recently, the practice of domain adaptation has seen the emergence of Adapter-based techniques, initially proposed by \citet{houlsby2019parameter}. Adapters are lightweight modules commonly used in both NLP and Speech to adapt pre-trained models to new tasks or domains. In speech-related tasks, Adapter-based fine-tuning has found applications in speech translation~\citep{le2021lightweight,gow2023naver,antonios2022findings}, domain adaptation~\citep{thomas2022efficient,tomanek2021residual}, and other tasks. They have demonstrated comparable performance to standard fine-tuning while utilizing only a fraction of trainable parameters.

Furthermore, there are efforts to adapt Whisper models to specific tasks using LoRA adapters (e.g. Arabic dialect identification~\citep{radhakrishnan2023parameter}, spoken language understanding~\citep{wang23ga_interspeech}, emotion recognition~\citep{feng2023peft}). This technique is elaborated in Section \ref{subsec:lora}. Additionally, some work involves full fine-tuning for task adaptation (e.g child spoken language understanding ~\citep{jain2023adaptation}).

In contrast to adapters and full fine-tuning, our work introduces gated Language-specific layers into the Whisper model and presents a parameter-efficient Knowledge Distillation approach. These innovations enhance the model's robustness to out-of-domain data.

\subsection{Low-rank Adapters (LoRA)} \label{subsec:lora}

Low-rank Adapter (LoRA) fine-tuning, as proposed by \citet{hu2022lora}, is a technique designed to reduce memory requirements for domain adaptation. This is achieved by introducing new trainable parameters into a pre-trained neural network while keeping the original pre-trained model weights fixed. These introduced parameters take the form of trainable rank decomposition matrices, and they are inserted between specific layers or blocks of the model. This approach significantly reduces the number of parameters that need to be fine-tuned when adapting the model for specific downstream tasks. For example, when fine-tuning a multilingual multi-task model for a single language and task, LoRA adapters help streamline the adaptation process.

The key assumption behind LoRA is that weight matrix updates in Transformer-based models exhibit a low "intrinsic rank" when undergoing full fine-tuning. This means that a pre-trained weight matrix, denoted as $W_0\in \mathbb{R}^{d\times k}$, can be effectively represented using a low-rank matrix decomposition, denoted as $W_0+\Delta W=W_0+BA$, where $B \in \mathbb{R}^{d\times r}, A\in \mathbb{R}^{r\times k}$, and the rank $r \ll \min(d,k)$. Importantly, during LoRA fine-tuning, the $W_0$ part remains fixed (frozen) and does not receive gradient updates, while $A$ and $B$ become sets of trainable parameters.

\begin{equation} \label{eq:lora}
h = W_0 x + \Delta W x = W_0 x + BA x
\end{equation}

One significant advantage of this approach is that it allows for parallel computation during the forward pass. Specifically, the forward pass output $h$ can be efficiently computed in parallel, and then the partial results are summed coordinate-wise, as presented in Equation \ref{eq:lora}.

\section{Quantization} \label{sec:quantization}

Quantization is a well-established technique in the field of Deep Learning, employed to increase the efficiency of neural networks. Historically, neural networks were often trained using low-precision numerical representations \citep{hubara2017quantized}. However, a recent trend, particularly in NLP, involves post-training quantization. This technique entails applying quantization to models after they have been trained with regular, higher precision. This approach has gained traction as it offers the dual benefits of reducing inference latency and model size.

Post-training quantization has found widespread use in various domains, including machine translation and language models \citep{wu2020integer, menghani2023efficient, liang2021pruning, bondarenko-etal-2021-understanding}. Quantized NLP models have yielded promising results, making it an appealing approach.

One of the most widely adopted techniques for post-training quantization in both NLP and speech communities is the \textbf{LLM.int8()} algorithm \citep{dettmers2022}. This method implements quantization in the feed-forward and attention projection layers of the Transformer architecture. The method has two parts: vector-wise quantization and mixed precision decomposition. In the vector-wise quantization, it is determined conversion constants that allow for the recovery of original numbers from 8-bit to 16-bit floating-point representations. This enables matrix multiplication to be carried out in the lower 8-bit precision. Moreover, in the mixed precision decomposition, it identifies potential outliers that could be adversely impacted by reduced precision and then executes this part of the matrix multiplication in 16-bit precision.

While initially designed for decoder-only large language models (LLMs), this quantization method, along with its 4-bit variation \citep{dettmers2023case}, has gained widespread adoption for various Transformer-based models. It has been made readily available in the Transformers library by Hugging Face ~\citep{wolf2020transformers}, contributing to its popularity. Additionally, it is becoming common to combine this quantization technique with domain adaptation methods. For instance, the QLoRA \citep{dettmers2023qlora} method incorporates LoRA adapters on top of a quantized Transformer model.

\section{Knowledge Distillation} \label{sec:kd-background}

Knowledge distillation~(KD) has been initially proposed by~\citet{kd_hinton} to distill knowledge from ensemble of models into a single model. Over time, KD has evolved to distill knowledge from a large teacher model into smaller student models \citep{sanh2020distilbert,mohammadshahi-etal-2022-small,shen2023language}. Knowledge distillation can be approached in two primary ways: representation matching or distribution matching. In this work, our focus is on distribution matching.

Traditional distribution matching knowledge distillation methods involves minimizing the Kullback–Leibler (KL) divergence between a teacher model and a student model. This is mathematically represented by Equation \ref{eq:KLdiv}:

\begin{equation} \label{eq:KLdiv}
J_{\text{KL}} = D_\text{KL}(p\|q_\theta)=\mathbb E_{\mathbf Y\sim p} \left[ \log \frac{p(\mathbf Y)}{q_\theta (\mathbf Y)} \right]
\end{equation}

where $p$ is the teacher distribution, $q_\theta$ is the student distribution, and $\mathbf Y$ is sampled from the teacher distribution.

However, learning based on KL divergence at the sequence level can often lead to the student distribution becoming overly smooth, as it attempts to cover the entire support of the teacher distribution. This behavior arises due to the asymmetric nature of the KL divergence, a phenomenon sometimes referred to as the \textit{mode-averaging problem}, as demonstrated by \citep{js_vs_kd_acl23}.

Recent research~\citep{js_vs_kd_acl23,js_vs_kd_bb2023} have shown that symmetric divergences, such as the Jensen-Shannon (JS) divergence, exhibit fewer borderline behaviors and tend to yield improved results in sequence-level distillation. Traditional JS divergence is expressed in Equation \ref{eq:JSdiv}:

\begin{equation} \label{eq:JSdiv}
    J_{\text{JS}} = D_\text{JS}(p\|q_\theta) = \frac12 \mathbb{E}_{\mathbf Y \sim p} \left[ \log \tfrac{p(\mathbf Y)}{m(\mathbf Y)} \right] + \frac12 \mathbb{E}_{\mathbf Y' \sim q_\theta} \left[ \log \tfrac{q_\theta (\mathbf Y')}{m(\mathbf Y')} \right]
\end{equation}

where $p$ is the teacher distribution, $q_\theta$ is the student distribution, $\mathbf Y$ and $\mathbf Y'$ are sampled from the teacher's and student's distributions and compared with their average $m(\cdot)=\frac12 p(\cdot) + \frac12q_\theta(\cdot)$.

\section{Datasets for Multilingual ASR} \label{sec:dataset-background}

Here we present two widely used massively-multilingual datasets that will be used in this work: CommonVoice~13.0 and FLEURS.

\subsection{CommonVoice~13.0}

The CommonVoice 13.0 (CV-13) corpus \citep{ardila-etal-2020-common}, represents the latest iteration of a massively multilingual collection of transcribed speech. It serves as a valuable resource for research and development in the field of speech technology. While primarily designed for Automatic Speech Recognition (ASR) applications, this dataset also finds utility in other domains, such as language identification. The utterances comprising this dataset are sourced from Wikipedia articles and supplemented with utterances contributed by language communities. These are subsequently narrated by contributors through Mozilla's website or iPhone app. To ensure data quality, contributions undergo validation by other volunteers, with only validated data being incorporated into the train, validation, and test subsets splits of the dataset. As of the current version, the dataset encompasses a rich tapestry of 110 languages, though the number of utterances per language varies significantly.

\subsection{FLEURS}

The FLEURS~\citep{conneau2023fleurs} is an n-way parallel speech dataset in 102 languages built on top of the machine translation FLoRes-101 benchmark \citep{goyal2022flores}, with approximately 12 hours of speech supervision per language. It was meant for few-shot learning on a variety of speech tasks, including Automatic Speech Recognition, Speech Language Identification, Speech Translation and Retrieval. The creation of this dataset involved the recording of all the publicly available sentences from FLoRes-101 (from dev and devtest split subsets). Each sentence was recorded by three paid native-speaker experts per language. Subsequently, these spoken sentences underwent a thorough evaluation by paid evaluators to ensure the overall quality and accuracy of the recorded content. The dataset is unbalanced as not all the sentences were validated, but most part of the languages have between 2400 and 3300 utterances on the train split, with an average 12 seconds per audio sample.

\section{The Whisper Model}
\label{sec:Whisper}

In this section we present Whisper \citep{radford2023robust}, the base model for the studies conducted in this work. 

\begin{figure}[caption={The Whisper model architecture (Source: \citet{radford2023robust})},label=fig:whisper-approach]
\includegraphics[width=1.0\textwidth]{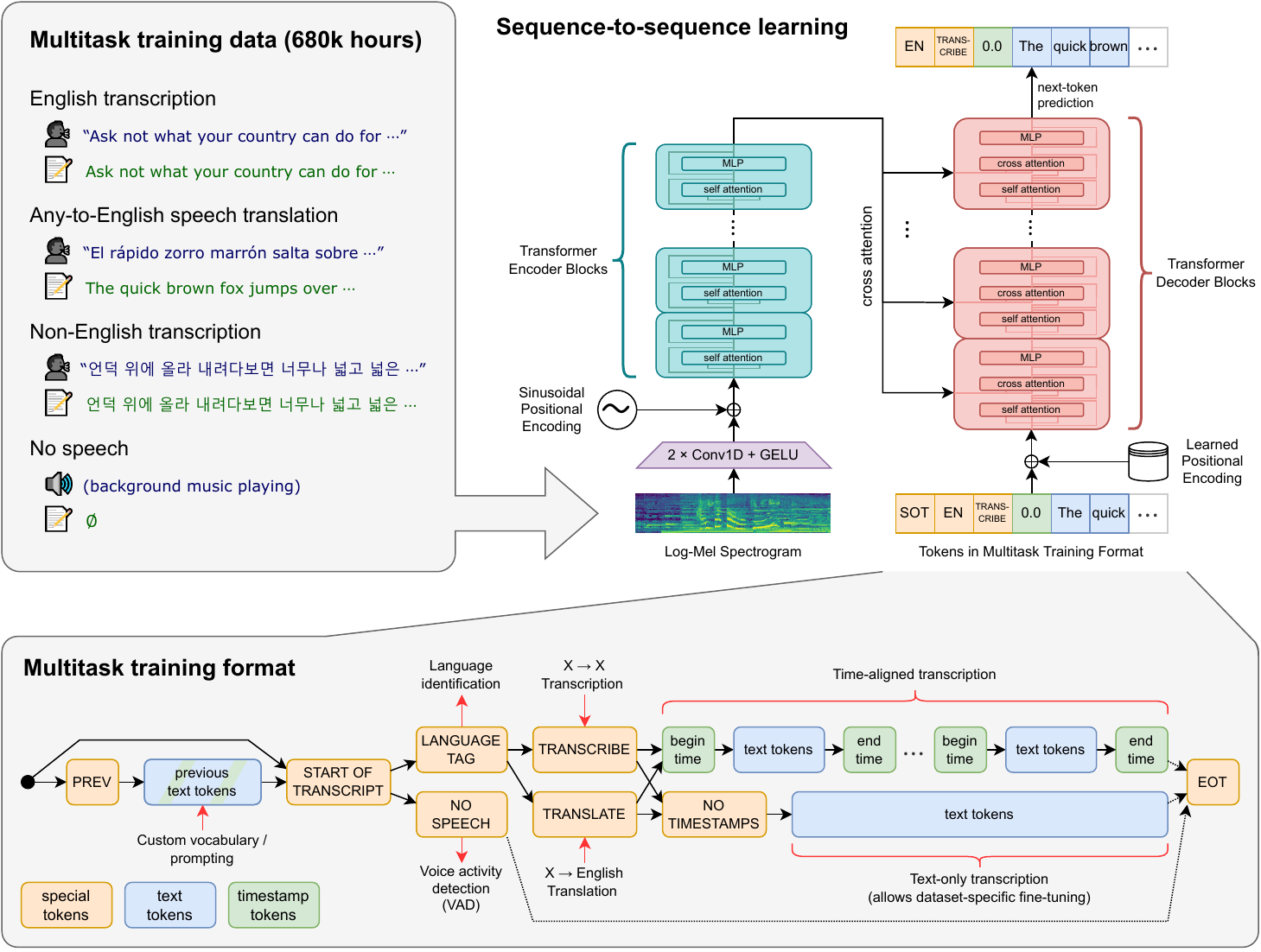}
\end{figure}

\subsection{Overview}

Whisper is designed to serve as a versatile end-to-end Automatic Speech Recognition (ASR) model suitable for a wide range of applications and languages. When it comes to ASR, previous research has predominantly focused on two key approaches: large-scale Unsupervised Learning \citep{wang2022wav2vec} and Supervised Learning as discussed in Section \ref{sec:sota-asr}. 

In the case of large-scale Unsupervised Learning, models benefit from training on vast, low-cost, and unlabeled datasets, which helps in building a high-quality encoding component. However, these models generate output that is not directly usable for ASR applications and requires further fine-tuning. On the other hand, Supervised Learning approaches utilize pretrained models that can be directly used for ASR tasks. However, they often struggle to generalize when faced with shifts in the data distribution, primarily due to the limited size of the datasets they were originally trained on. Additionally, creating large-scale human labeled datasets for these models can be prohibitively expensive.

Whisper takes a unique approach by introducing Weakly Supervised Learning, striking a balance between data quality and quantity. The Whisper training dataset is curated by collecting pairs of audio and corresponding transcripts from the internet (mainly YouTube videos). After some minimal processing, that included employing language identification with the model proposed by \citet{valk2020}, this dataset comprises a substantial $680,000$ hours of highly diverse audio content. Notably, it encompasses 96 languages besides English, with approximately 17.2\% of the dataset consisting of audio and transcript pairs in the same language (ASR). Additionally, around 18.4\% of the pairs have English-translated transcripts.

This unique approach provides Whisper with several advantages. Firstly, the Whisper encoder benefits from the rich and diverse dataset, making it perform exceptionally well, similar to Unsupervised settings. Secondly, Whisper is trained with relatively clean labels, allowing it to be used in a Zero-Shot manner without the need for extensive fine-tuning.

\subsection{Architecture}
\label{subsec:whisper_architecture}
The architecture of Whisper consists of the original Transformer architecture~\citep{vaswani2017attention} preceded by dimension reduction layer called stem. The architecture is visually depicted in Figure~\ref{fig:whisper-approach}.

\subsubsection{Stem} 

The stem comprises a pair of 1-dimensional Convolution Layers, each accompanied by GELU activations. Both convolution layers employ filters of size $3$ and produce $d$ output channels. The value of $d$ varies across different sizes of the Whisper architectures. The first convolution layer operates with a stride of $1$, while the second employs a stride of $2$ (effectively reducing the length of the input sequence by half). Consequently, the output of the stem consists of a sequence of $1500$ elements, each with dimension $d$. As the self-attention layers in a Transformer exhibit quadratic complexity concerning the sequence length, for a fixed hidden representation size of $d$, the stem significantly reduces the computational complexity by a factor of $4$.

\subsubsection{Transformer}

In their work, \citet{radford2023robust} primarily highlights the impact of scaled Weak Supervision on ASR system performance, with less emphasis on architectural modifications. The base architecture employed for Whisper is the encoder-decoder Transformer, which is renowned for its scalability and reliability in several sequence-to-sequence tasks.

However, the Whisper Transformer does introduce a few key modifications compared to the original Transformer architecture. Sinusoidal encodings are added to the input representations of the encoder, while the positional encodings in the decoder are learned. Additionally, GELU activation functions are used instead of ReLU, and these activations are applied following the residual blocks. Moreover, a normalization layer is included in the encoder's output. Furthermore, Whisper offers a range of five different architecture sizes, as detailed in Table~\ref{tab:whisper-sizes}. These varying sizes cater to different requirements and performance needs, allowing for flexibility in ASR tasks.

\begin{table}[caption={Architectural specifications for the Whisper model family. $L$ denotes the number of layers per block, indicating that, for example, the tiny model with $L=4$ consists of 4 transformer layers in the encoder and 4 in the decoder.},label=tab:whisper-sizes]
  \centering
  \begin{tabular}{@{}l|ccc@{}}
    \toprule
    Model & Layer ($L$) & Width ($d$) & Parameters \\
    \midrule
    Tiny & 4 & 384 & 39M \\ 
    Base & 6 & 512 & 74M \\ 
    Small & 12 & 768 & 244M \\ 
    Medium & 24 & 1024 & 769M \\
    Large & 32 & 1280 & 1550M \\ 
    \bottomrule
  \end{tabular}
\end{table}

\subsubsection{Tokenization}
To tokenize transcripts, the Whisper model employs the BPE (Byte Pair Encoding) tokenizer originally introduced in GPT-2 by \citet{radford2019language}. When dealing with languages other than English, the tokenizer is adapted by refining it until the vocabulary size matches that of English. 

\subsection{Multitasking}

Whisper is trained and operates as a multitask model, capable of handling various sub-tasks within a single end-to-end architecture. These sub-tasks encompass Voice Activity Detection, Language Identification, Text Alignment, Transcription, Translation, and more. To delineate each task and the expected format of the subsequent predictions, specific tokens are employed, as delineated in Table~\ref{tab:special-tokens}. These tokens are positioned at the start of the output sequence, providing task context (see Figure \ref{fig:whisper-approach}). Token generation follows an auto-regressive process, reliant on prior tokens. For example, when the detected language is French, the model computes the likelihood of token $w$ at position $k'$, as illustrated in Equation~\ref{eq:conditional-prob}:

\begin{equation} \label{eq:conditional-prob}
P(w_{k'}=w|\dots,\texttt{<|fr|>},\texttt{|transcribe|},\dots,w_{k'-1},X)
\end{equation}

Consequently, the generated tokens will probably only belong to the French vocabulary as they have higher conditional probabilities compared to ones belonging to other languages. 

\begin{table}[caption={Subset of special tokens associated with Whisper's multitasks. For Language Identification, each language is specified with a token, and a single token is added to the sequence. This token is required. For Voice Activity Detection, only when the audio does not contain clear speech that its corresponding token is present in the output. The tasks Transcribe and Translate are mutually exclusive, but one of them is required.},label=tab:special-tokens]
    \centering
    \begin{tabular}{@{}m{4.5cm}|m{4.4cm}@{}}
        \toprule
        \textbf{Tasks} & \textbf{Tokens}  \\ \midrule
        Language Identification & \texttt{<|LANGUAGE|>} \\ 
        & e.g. \texttt{<|en|>}, \texttt{<|gl|>}, \texttt{<|fr|>},  \texttt{<|fa|>}, etc. \\ \hline
        Voice Activity Detection & \texttt{<|nospeech|>} \\ \hline
        Transcribe & \texttt{<|transcribe|>} \\ \hline
        Translate & \texttt{<|translate|>} \\ \hline
        Alignment & \texttt{<|notimestamps|>} \\ \bottomrule        
    \end{tabular}
\end{table}

Additionally, certain special tokens can be predefined to simplify predictions. In our work, we specifically enforce transcription and language tokens, thereby eliminating dependency on Language Identification quality for under-represented languages. Tasks not pertinent to our study are also disregarded.

		\chapter{Bias Analysis on Quantized Speech Models}
\label{chap:bias}

In this chapter, we aim at addressing the two first objective of the internship: understand the bias presented on Whisper models, and investigate how these are impacted by the employment of quantization. 

\section{Experimental Setup}

\subsection{Dataset preparation}

In our research, we employed the two widely recognized datasets described in Section \ref{sec:dataset-background}: FLEURS and Common Voice 13.0 (CV-13). These datasets provide valuable speaker-related information, including gender, language group (in the case of FLEURS), accent (exclusive to CV-13), and age (exclusive to CV-13).

Building upon the information available in FLEURS, we curated a gender-balanced benchmark, which we refer to as \textbf{Balanced-FLEURS}. The primary goal here was to mitigate the influence of confusion variables such as sentence complexity and gender imbalance (where certain languages exhibit a higher percentage of speakers from one gender). To achieve this, we mixture the train, validation, and test sets of FLEURS, meticulously filtering them to ensure that each sentence was narrated by both a male and a female speaker. Meanwhile, we also ran a Voice Activity Detection model on the dataset, as we encountered a notable number of empty audio files in Spanish, Norwegian, and Malay\footnote{We have reported this issue to the Google Team via HuggingFace, listing all problematic files. The corresponding issue can be found here: \\ \url{https://huggingface.co/datasets/google/fleurs/discussions/16\#6442a217f8b647fa4f50c489}}. We include in the experiments only the languages in which we were able to find at least 200 utterances. 

In addition to \textbf{Balanced-FLEURS}, we made use of the Common Voice 13.0 dataset, specifically its validation set. In this case, we leveraged gender and age information. While we attempted to incorporate accent information in our study as well, we encountered challenges in aggregating a sufficiently large dataset, even after merging the train, test, and validation splits. Consequently, we do not report our results with respect to accents.

\subsection{Resourcefulness categorization} \label{sec:resourcefulnessclass}

In the course of our experiments, we have introduced a resourcefulness classification system specifically tailored to weakly-supervised speech models, with a primary focus on the transcription task (ASR). This categorization is designed to group languages based on the amount of training data used in the model pre-training. The classification involves clustering languages into categories with similar amounts of training data, and the intervals used for this classification can be found in Table~\ref{tab:resourcefulnessclass}. 

\begin{table}[caption={Proposed Language resourcefulness categorization for Weakly-supervised ASR models},label=tab:resourcefulnessclass]
\begin{tabular}{@{}cc@{}}
\toprule
\multirow{2}{*}{\textbf{Resourcefulness}} & \multirow{2}{*}{\textbf{ASR Training data (h)}} \\
                                          &                                                 \\ \midrule
Super High-Resource                       & $\geq$ 5000                               \\
High-resource                             & {[}1000, 5000)                                  \\
Mid-to-high-resource                      & {[}500, 1000)                                   \\
Low-to-mid-resource                       & {[}100, 500)                                    \\
Low-resource                              & {[}10, 100)                                     \\
Extremely Low-Resource                    & (0, 10)                                         \\ \bottomrule
\end{tabular}
\end{table}

It is worth noting that our proposed classification system has a limitation in the context of Whisper. Specifically, it does not account the volume of training data available for the speech translation task. While this data does not directly impact the quality of generated text data for a language (since in Whisper, translation data available is to English only), it does play a role in enhancing the model's speech encoding capabilities.

\section{Bias evaluation on Whisper}

In this section, we present preliminary experiments conducted on the Whisper model. Our aim here is to investigate whether bias exists in the original versions of Whisper. To achieve this, we compare Whisper's performance on the validation split of CV-13 and on Balanced-FLEURS. Our analysis involves an aggregate approach, where we average the metrics across languages.

\begin{figure}[caption={Performance across languages on whisper-large-v2 on Balanced-FLEURS. Languages are ranked on x-axis based its performance.},label=fig:fleurs1.1]
\centering
\includegraphics[width=\linewidth]{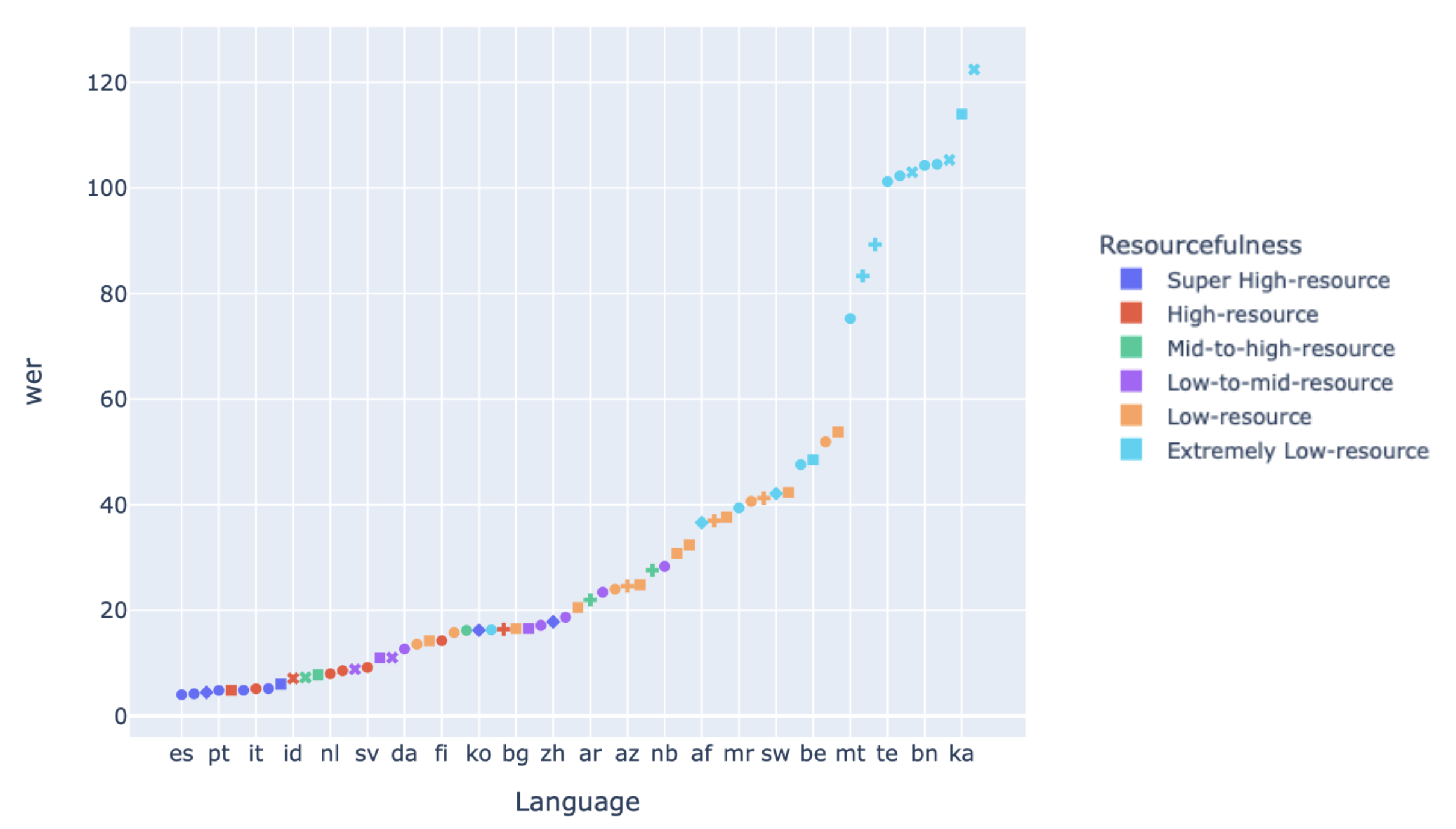}
\end{figure}

\begin{figure}[caption={Performance across languages on whisper-large-v2 on CV-13. Languages are ranked on x-axis based its performance.},label=fig:cv1.1]
\centering
\includegraphics[width=\linewidth]{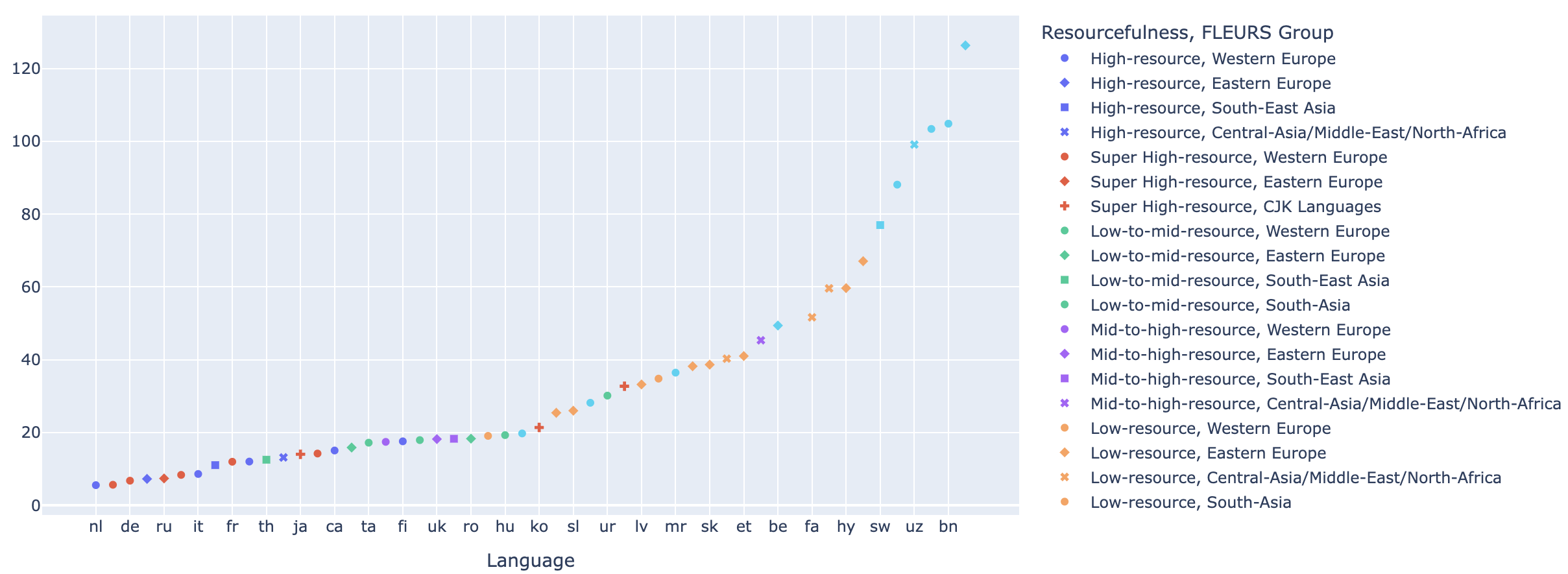}
\end{figure}

Figures \ref{fig:fleurs1.1} (Balanced-FLEURS) and \ref{fig:cv1.1} (CV-13) showcase the Word Error Rate (WER) performance across the languages covered in the two datasets for \texttt{whisper-large-v2}. These results reveal a clear correlation between performance and resourcefulness, with lower resource languages (Low and Extremely Low-Resource) consistently exhibiting poorest performance. Naturally, the impact varies among languages, possibly due to their complexity or the amount of training data available for closely-related languages. These findings collectively suggest a bias linked to resourcefulness.

\begin{figure}[caption={Average relative WER difference between male and female voice for Balanced-FLEURS. Languages are ranked on x-axis based its relative difference and resourcefulness.},label=fig:fleurs1.2]
\centering
\includegraphics[width=\linewidth]{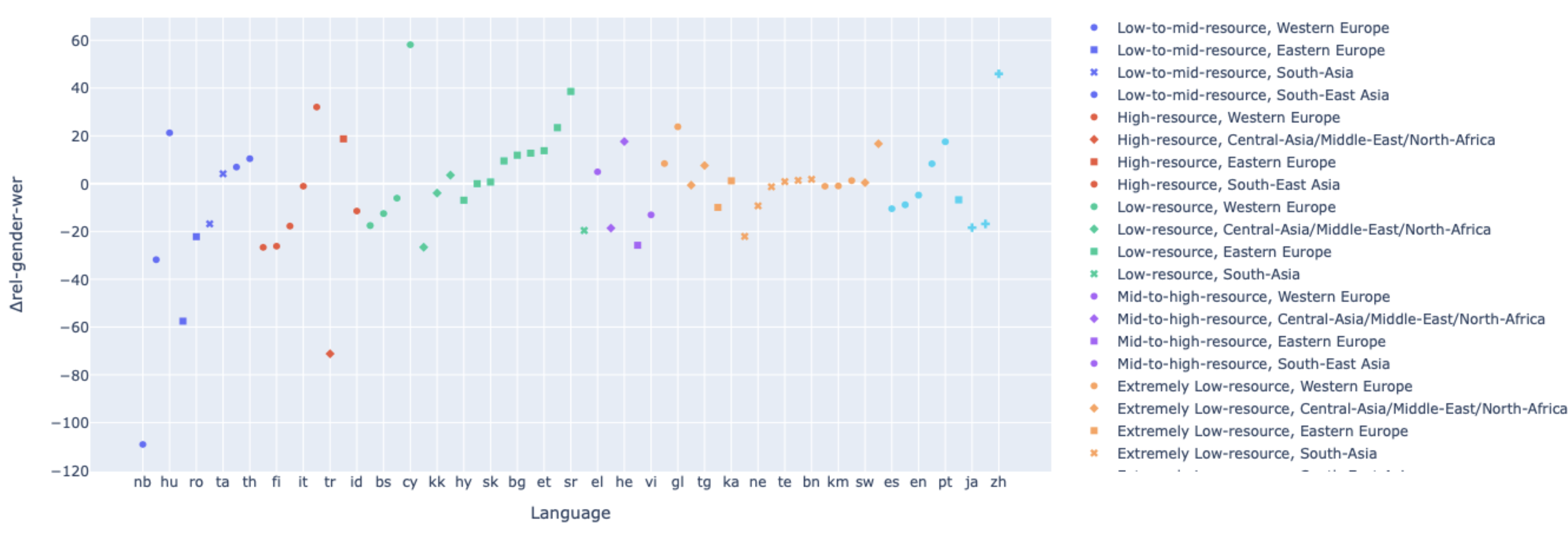}
\end{figure}

\begin{figure}[caption={Absolute WER difference between male and female voice for CV-13. Languages are ranked on x-axis based its absolute difference.},label=fig:cv1.2]
\centering
\includegraphics[width=\linewidth]{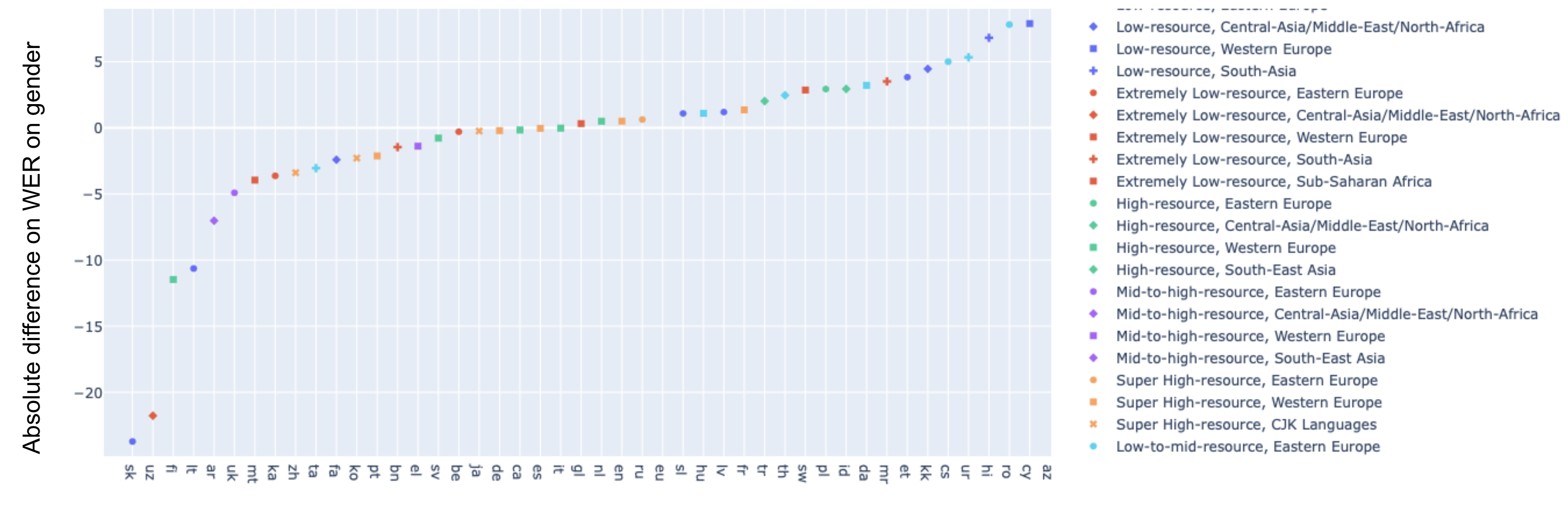}
\end{figure}

Figure \ref{fig:fleurs1.2} illustrates the average relative difference between male and female speakers for Balanced-FLEURS on \texttt{whisper-large-v2}. This metric, already employed is previous similar study by \citet{boito2022study}, is relevant here as the sentences are consistently the same across genders. Meanwhile, Figure \ref{fig:cv1.2} displays the absolute difference (following \citet{costa2022evaluating}) in WER between male and female speakers on CV-13. In both cases, the results show varying degrees of gender bias across different languages. Remarkably, these biases are consistent across the different datasets, implying that each language possesses its unique bias, likely attributed to the quality and diversity of its training data. While the model does exhibit gender bias, it is essential to note that, for the most part, this bias remains within a maximum average WER difference of 3 for the majority of languages (in the case of CV-13). 

\begin{figure}[caption={Performance across languages and across different whisper sizes on Balanced-FLEURS. Languages are ranked on x-axis based its resourcefulness.},label=fig:fleurs1.3]
\centering
\includegraphics[width=\linewidth]{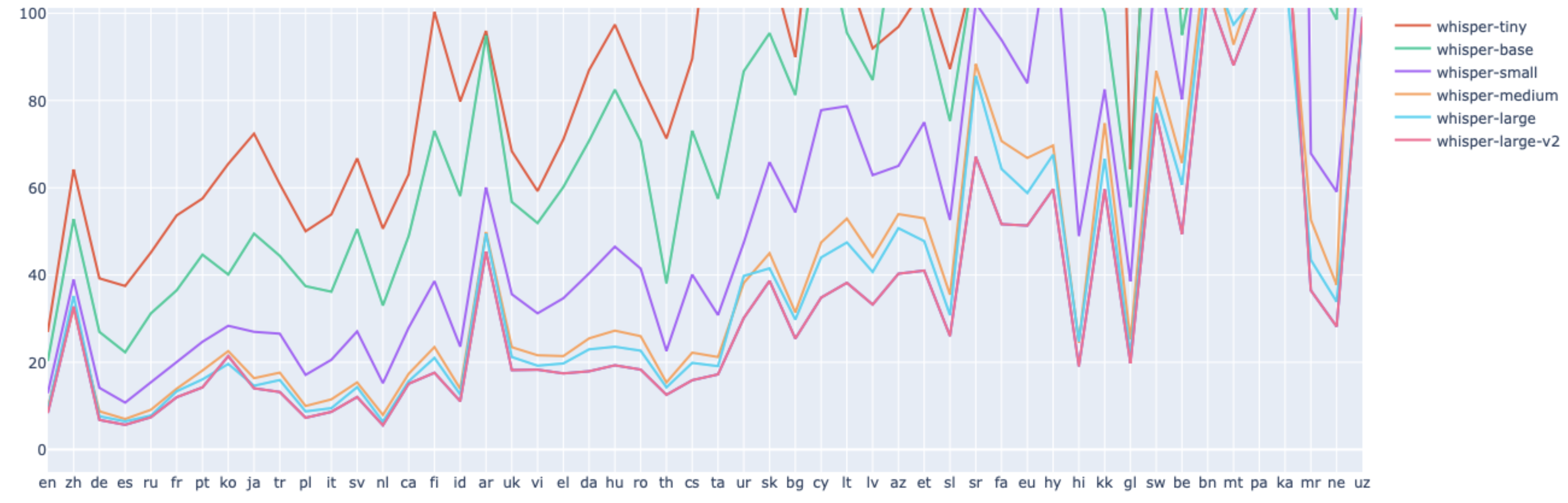}
\end{figure}

Figure \ref{fig:fleurs1.3} extends the analysis by presenting WER performance across different languages on Balanced-FLEURS, mirroring Figure \ref{fig:fleurs1.1}. However, this time, we consider all available model sizes within the Whisper family. Languages are ranked by resourcefulness. These results unveil two significant findings: (i) the performance trend aligns across nearly all languages, suggesting a consistent ranking of languages based on performance across all models; and (ii) notably, a clear correlation emerges between smaller model sizes and reduced performance, with the model curves closely overlapping. This phenomenon likely stems from the \textit{curse of multilinguality}, wherein less resourceful languages exhibit larger performance disparities among model sizes. Additionally, it's worth noting the differences between large and large-v2 models. Although both models share the same size, the former benefits from more extensive training, additional optimization steps, and data augmentation techniques. Finally, these findings collectively shed light on bias associated with architecture size, despite models being trained with the same dataset.

\section{Bias evaluation on quantized Whisper}

Now, we delve into the quantized version of Whisper. In this set of experiments, we apply the \texttt{LLM.int8()} method \citep{dettmers2022} (described in Section \ref{sec:quantization}) to Whisper. The primary objective of this study is to investigate whether the biases observed in the original Whisper model persist, diminish, or intensify after quantization. In essence, we seek to understand what model features may be forgotten due to quantization.

In contrast to the previous section, our analysis here adopts a sentence-level approach. We compare the model's performance on individual sentences before and after quantization. To ensure a fair evaluation, we exclude sentences with initial Word Error Rate (WER) values greater than or equal to 100. For this sentence-level analysis, we create histograms based on the absolute difference in WER before and after compression. We categorize sentences into three groups: those that worsened (WER increased by more than 5), those that remained similar (WER difference less than 5), and those that improved (WER reduced by more than 5).

\begin{figure}[caption={Histogram of performance degradation by quantization per gender on Balanced-FLEURS},label=fig:fleurs2.1]
\centering
\includegraphics[width=\linewidth]{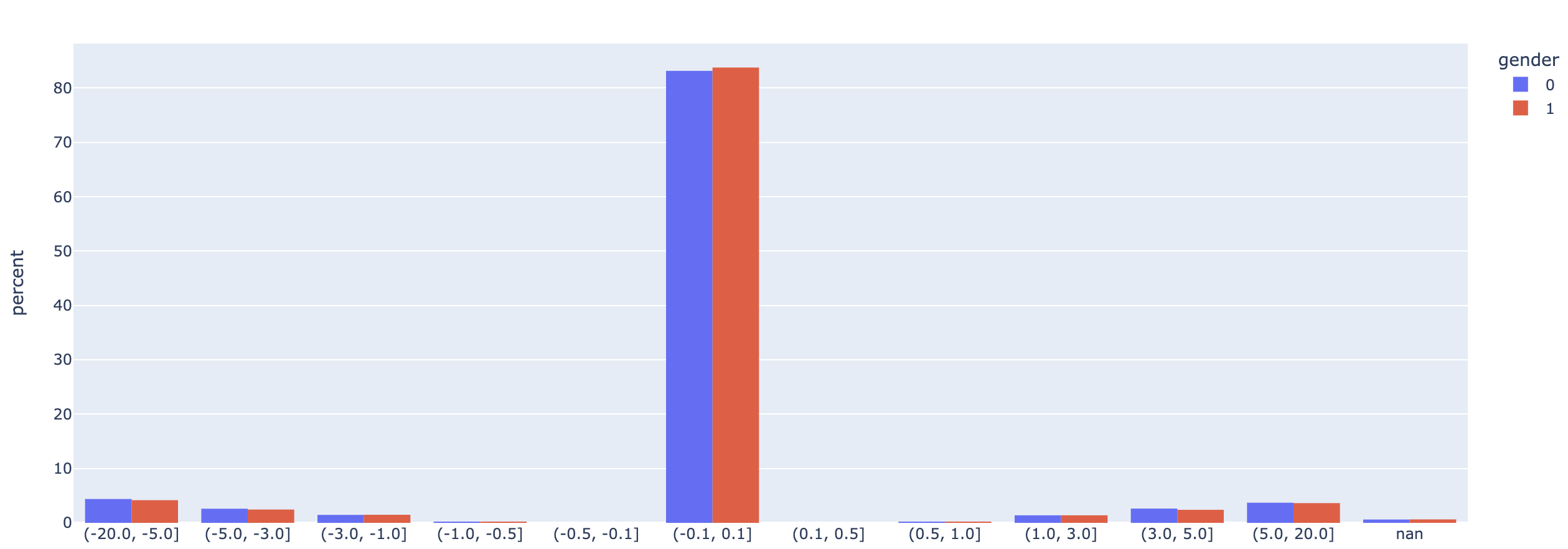}
\end{figure}

\begin{figure}[caption={Histogram of performance degradation by quantization per gender on CV-13},label=fig:cv2.1]
\centering
\includegraphics[width=\linewidth]{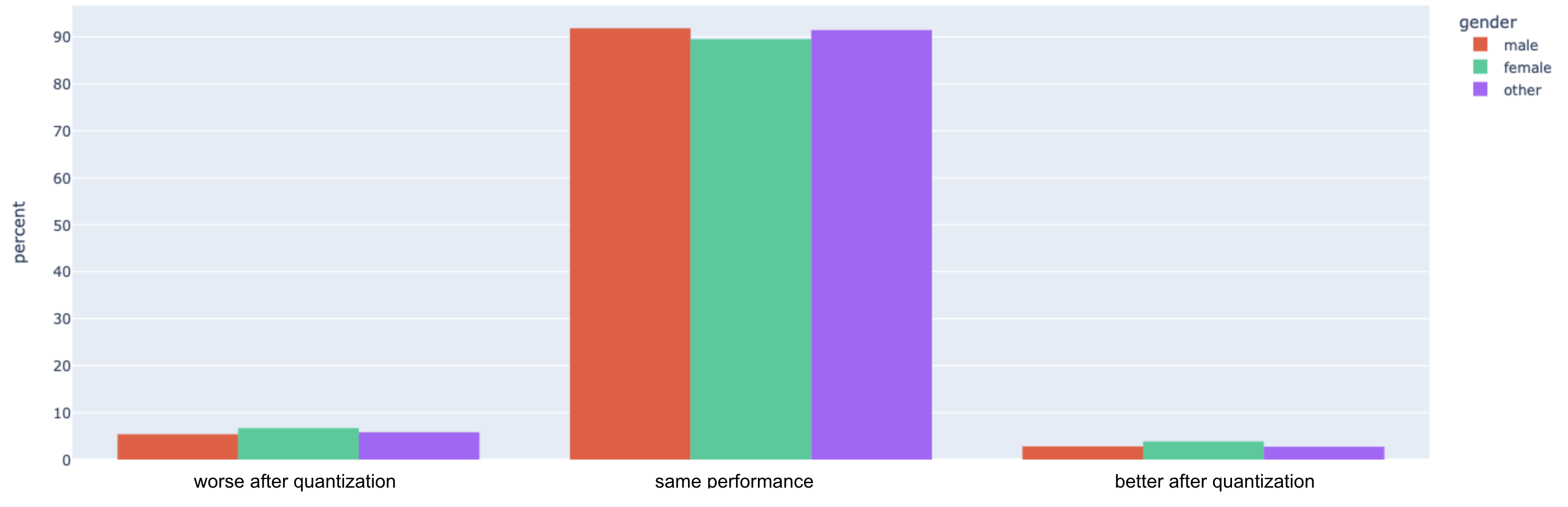}
\end{figure}

\begin{figure}[caption={Histogram of performance degradation by quantization per age group on CV-13},label=fig:cv2.2]
\centering
\includegraphics[width=\linewidth]{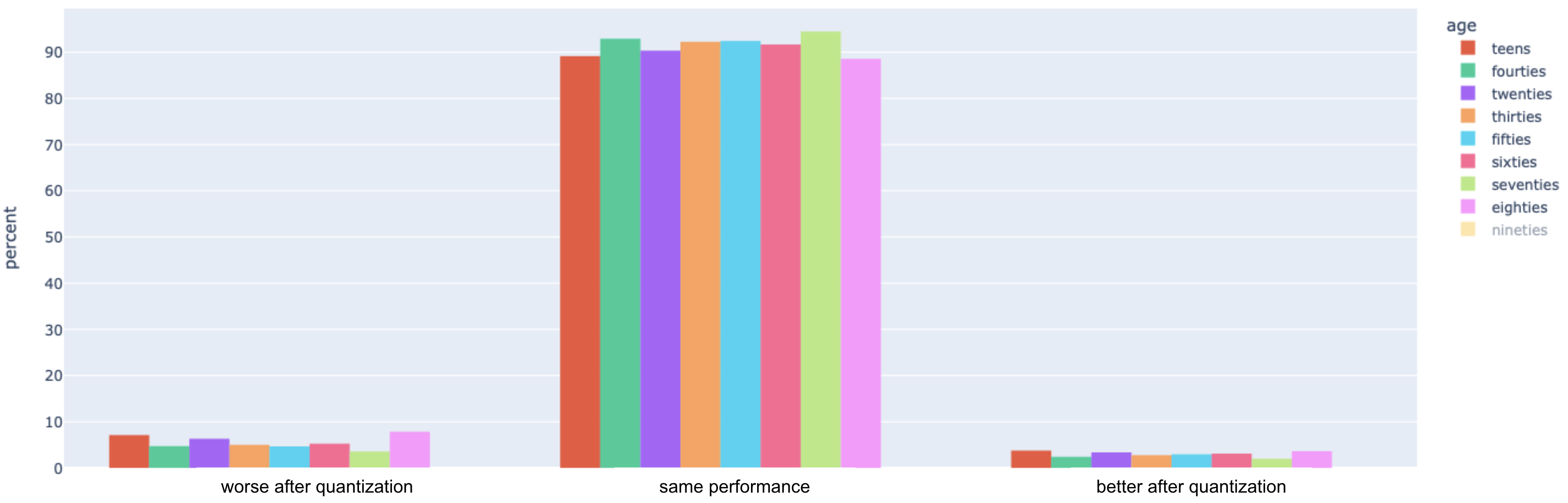}
\end{figure}

Figures \ref{fig:fleurs2.1} (Balanced-FLEURS) and \ref{fig:cv2.1} (CV-13) present histograms categorized by gender for the \texttt{whisper-large-v2} model. Figure \ref{fig:cv1.1} displays histograms categorized by age group for CV-13. The data clearly indicates that quantization equally impacts all genders and age groups, implying that gender and age biases are kept unchanged after quantization.

\begin{figure}[caption={Histogram of performance degradation by quantization per resourcefulness group on Balanced-FLEURS},label=fig:fleurs2.2]
\centering
\includegraphics[width=\linewidth]{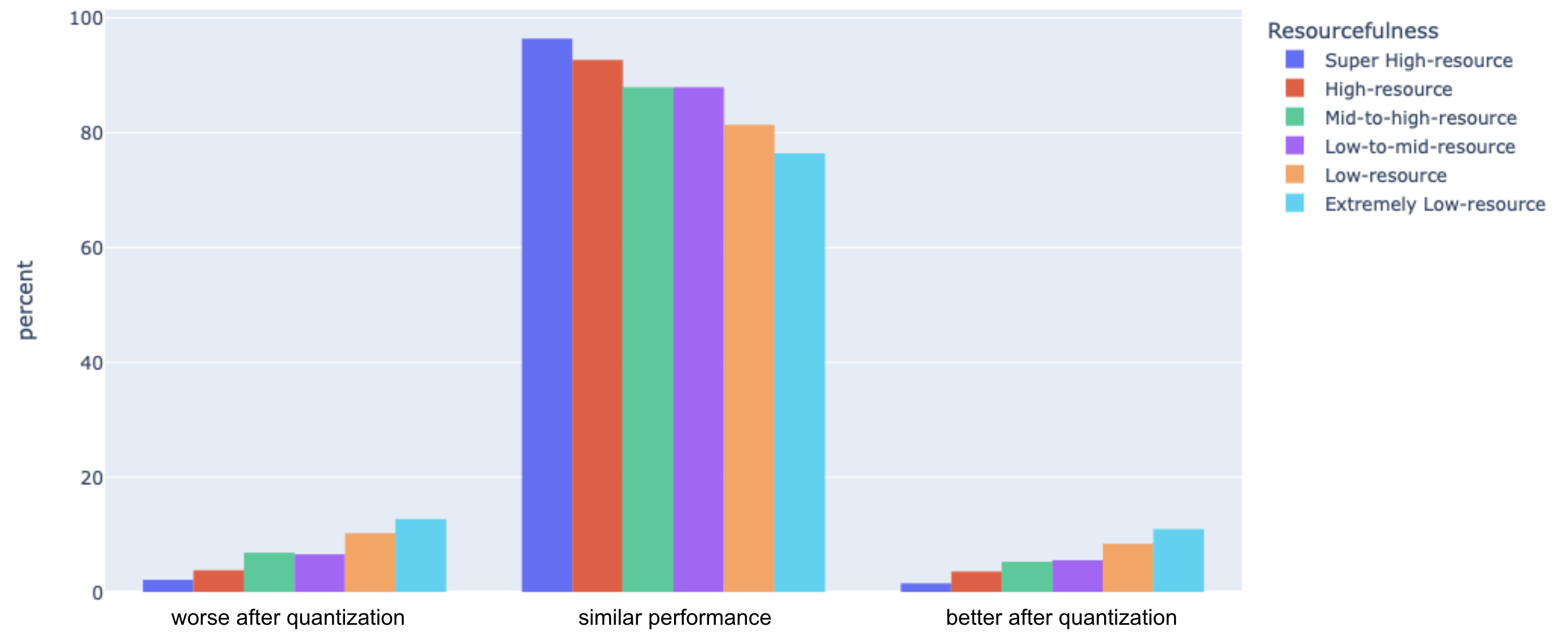}
\end{figure}

\begin{figure}[caption={Histogram of performance degradation by quantization per resourcefulness group on CV-13},label=fig:cv2.3]
\centering
\includegraphics[width=\linewidth]{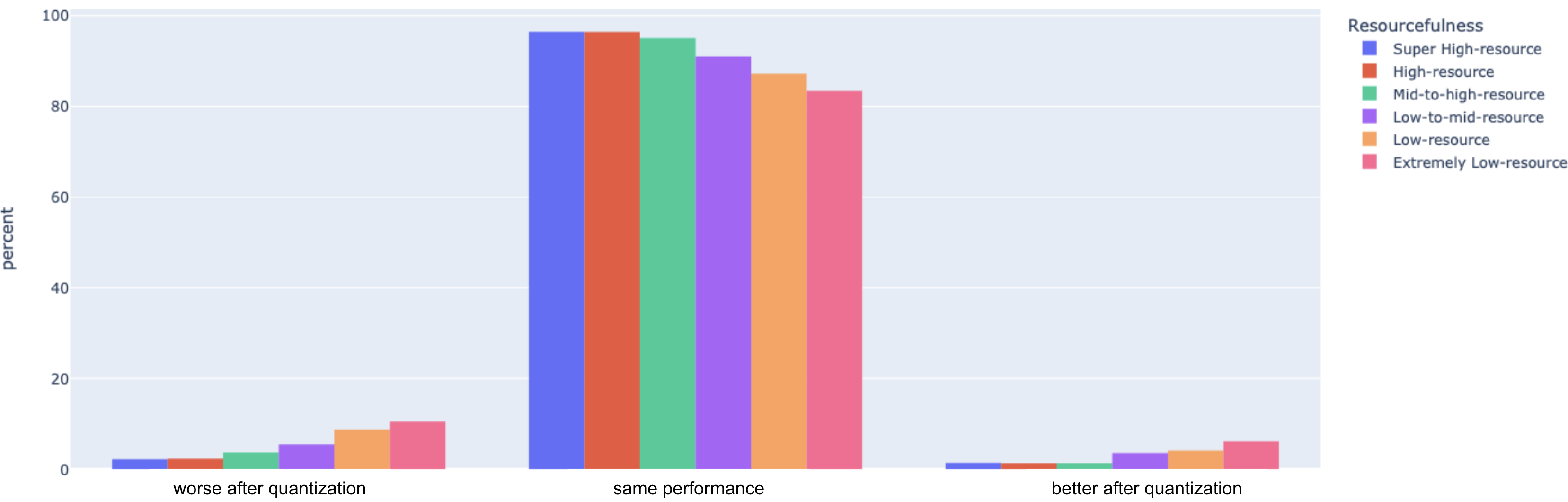}
\end{figure}

In figures \ref{fig:fleurs2.2} (Balanced-FLEURS) and \ref{fig:cv2.3} (CV-13), we illustrate histograms categorized by language resourcefulness for \texttt{whisper-large-v2}. Here, a distinct pattern emerges: lower-resource languages are more significantly affected by quantization. While almost all sentences in super high-resource languages maintain their performance, approximately 25\% of sentences in extremely low-resource languages are impacted (in the case of Balanced-FLEURS). Consequently, quantization amplifies the resourcefulness bias.

\begin{figure}[caption={Histogram of performance degradation by quantization per model size on Balanced-FLEURS},label=fig:fleurs2.3]
\centering
\includegraphics[width=\linewidth]{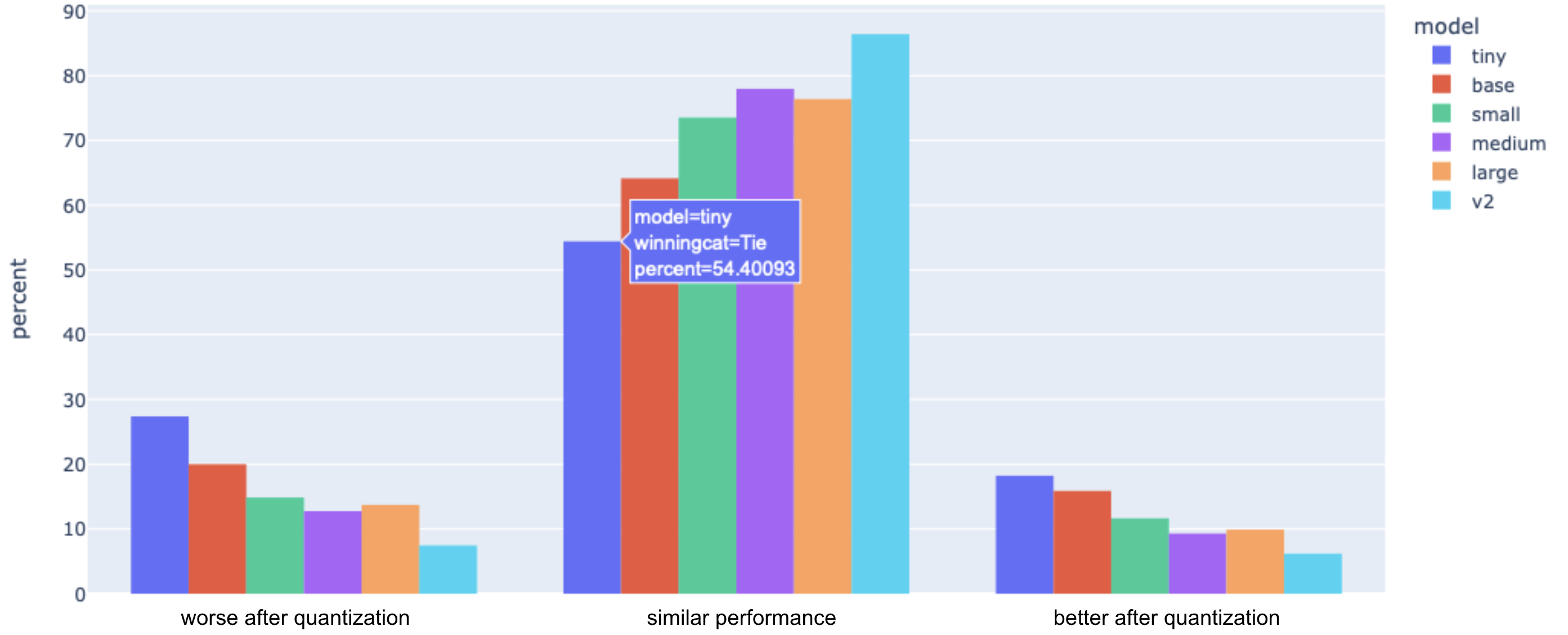}
\end{figure}

Lastly, in figure \ref{fig:fleurs2.3} (Balanced-FLEURS) and \ref{fig:cv2.4} (CV-13), we present histograms considering all available model sizes within the Whisper family, grouped by model size. The results highlight significant differences in how quantization affects models of varying sizes. While a small proportion of sentences are impacted for \texttt{whisper-large-v2}, there is a striking contrast, with almost half of the sentences affected in the case of \texttt{whisper-tiny}. This highlights that the bias related to architecture size is significantly amplified by quantization.

This last finding indicates that smaller models are generally more susceptible to the effects of quantization. This observation is particularly concerning as many parameter-efficient domain adaptation methods in use today in NLP and Speech involve applying quantization first, without considering the model size. This calls for practitioners to exercise caution when adapting pre-trained models to avoid the addition of unintended bias.


\section{Summary of the main findings}

Here we present the key takeaways from this chapter. First, Whisper exhibits certain speaker-related biases, such as gender and age. These biases are kept unchanged after applying quantization to the model.

On the other hand, biases associated with the model itself (model-related bias), including language resourcefulness and architecture size, are amplified by quantization. Overall, Low-resource languages are the most adversely affected by quantization. Moreover, there is a clear pattern on the architecture size, with smaller models experiencing more significant performance degradation compared to larger ones. This is concerning as current parameter-efficient approaches (such as QLoRA presented on Section \ref{sec:quantization}) mostly apply quantization first, regardless of the model size.

This presents a significant challenge: Can we enhance the performance of smaller models for languages where they currently perform poorly, even though the best model performs well? We aim at searching an alternative to quantization to reduce the model size.

            \chapter{DistilWhisper}
\label{chap:distilwhisper}



One prominent observation is the significant Automatic Speech Recognition (ASR) performance gap between the \texttt{whisper-large-v2} model and its counterparts of smaller sizes, especially when applied to a diverse set of languages. This gap in performance is noticeable across a wide spectrum of languages, that include the low-resource ones, but also many mid- and high-resource languages. As our earlier analysis, outlined in Chapter \ref{chap:bias}, revealed, the "lower" resource languages are also the most affected by lightweight compression techniques. 

This phenomenon is often referred to as the \textit{curse of multilinguality} (as discussed in related works by \citet{mmmt_google,conneau2020unsupervised, goyal2021largerscale}). It stems from the inherent challenge that arises when attempting to cover an extensive array of languages within a single model - the performance inevitably suffers unless the model is significantly scaled up. This leads us to the central question that has motivated our research: Can we improve the performance of smaller models for languages in which they currently perform poorly, but the best model performs well? 

A common approach to address this challenge of achieving efficient inference could be distilling knowledge from a larger multilingual teacher model into a smaller pre-existing one, as highlighted in prior works such as the ones done by \citet{sanh2020distilbert} and \citet{mohammadshahi-etal-2022-small}. However, when it comes to applying such knowledge distillation (KD) to \texttt{whisper-large-v2}, which represents the best and largest Whisper model, we face a significant hurdle. This is because we need access to information that is not readily available, such as comprehensive training data spanning all tasks and languages, and its original learning objective, in order to maintain the original model's robustness.

Recent research findings, exemplified by works like \citet{pfeiffer-etal-2022-lifting} and \citet{pratap2023scaling}, have demonstrated an alternative solution to the \textit{curse of multilinguality}. This approach involves equipping moderately sized models with language-specific (LS) modules. This sparse architectural design permits the extension of model parameters through additional modules as more languages are incorporated into the model. Consequently, it ensures consistent performance across languages without incurring substantial additional computational costs during inference.

In light of the overarching goal to enhance model performance for various languages within the constraints of limited model capacity, our work introduces the \textit{DistilWhisper} approach. We incorporate conditional language-specific routing (CLSR) modules, as described by \citet{zhang2021share}, into a smaller version of  Whisper. We then optimize these modules jointly through ASR fine-tuning and knowledge distillation from a larger Whisper model (\texttt{whisper-large-v2}). For a visual representation of our architecture, please refer to Figure \ref{fig:arch}, and in the subsequent sections, we delve into the key components of our approach.

Following, in this chapter, we detail the elements that make up our approach. Then, in the next chapter (Chapter \ref{chap:results}), we will present how we validate this approach and its results following the \textit{DistilWhisper} approach presented here.

\begin{figure*}
\centering
\includegraphics[width=\linewidth]{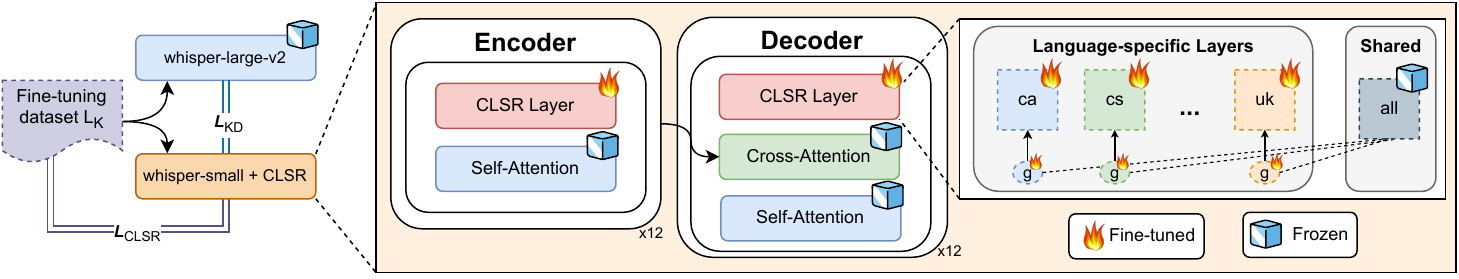}
    \caption{The \textit{DistilWhisper} optimization approach~(left), and its architecture~(right). The feed-forward is replaced by a CLSR module, where the LS gates~(g) learn to alternate between the pre-trained frozen multilingual representation and the LS layer.}
\label{fig:arch}
\end{figure*}


\section{Conditional Language-Specific Routing}

We extend Conditional Language-Specific Routing (CLSR) modules proposed by \citet{zhang2021share}, and commonly used in Multilingual Neural Machine Translation, for the first time to the speech domain. This module, which introduces sparsity to the Transformer architecture, learns a hard binary gate $g(\cdot)$ for each input token by using its hidden embedding $z^l$. These decisions enable a layer to selectively guide information through either a LS path denoted as $h^{lang}$ or a shared path referred to as $h^{shared}$, as in Eq.~\ref{eq:clsr}: 

\begin{equation}\label{eq:clsr}
     \text{\texttt{CLSR}}(z^l) = g(z^l)\cdot h^{lang}(z^l) + (1 - g(z^l))\cdot h^{shared}(z^l). 
\end{equation}

In contrast to the original CLSR, in this work we use language-specific gates as shown in Figure~\ref{fig:arch}, instead of sharing them across languages. This allows us to train language-specific components individually~(i.e. in parallel), and then only load the relevant modules at inference. Moreover, our approach also differs from the original CLSR by the positioning: supported by previous work~\citep{zhang2021share, pfeiffer-etal-2022-lifting}, we limit CLSR to the feed-forward network (correspondent to the feature domain of the Transformer architecture), which we also replace entirely by the CLSR module, reducing the increment in the number of parameters. 

Following the proposal from \citet{zhang2021share}, each gate $g(.)$ is made by a two-layer bottleneck network, which is summed to a increasing zero-mean Gaussian noise during training to discretize it: 
\begin{equation}\label{eq:gate1}
g(z^l) = \sigma(G(z^l) + \alpha(t)\cdot\mathcal{N}(0, 1))\text{,}
\end{equation}
\begin{equation}\label{eq:gate2}
\text{with }G(z^l) = \text{\texttt{ReLU}}(z^l W_1 + w_2)\text{,}
\end{equation}

where $\sigma (\cdot)$ is the logistic-sigmoid function, and $W_1$ and  $w_2$ are trainable parameters. $\alpha$ is linearly increased along with training steps $t$. At inference time, we adopt hard gating: 
\begin{equation}\label{eq:gate1}
g(z^l) = \delta (G(z^l) \ge 0)\text{,}
\end{equation} 
where $\delta (\cdot)$ is a Dirac measure.

\section{\textit{DistilWhisper} approach} 

Figure~\ref{fig:arch} presents our proposed \textit{DistilWhisper} architecture. 
Our student 
is enriched with CLSR modules at each feed-forward for each language. These all experts in each CLSR layer are equally initialized from the frozen weights of the corresponding feed-forward layer. At training time, for each language the model updates only the corresponding language-specific experts and gates. At inference time, the model loads the shared layers~(multilingual) and the Language-Specific experts and gates for the languages of interest, resulting in a limited parameter overhead. We highlight that the use of CLSR modules brings more flexibility to our architecture when compared to adapters, as it allows for routing at the token-level. This makes this approach more capable of leveraging pre-existing knowledge~(shared frozen module), activating the Language-Specific path only when this is likely to increase performance.

\section{\textit{DistilWhisper} optimization}

The optimization of our \textit{DistilWhisper} architecture consist of a standard cross-entropy loss, along with two new elements: gate budget loss, and knowledge distillation. Following we detail these new elements. 

\subsection{Gate budget loss}

Following \citet{zhang2021share}, when learning CLSR module parameters, in addition to standard cross-entropy loss $\mathcal{L}_{\text{CE}}$, we optimize a gate budget loss $\mathcal{L}_{\text{g}}$ to balance models' usage of language-specific and shared modules. It relies on the gate $g(.)$ activation values for a pair (audio, text) $(X,Y)$ in a batch $\mathcal{B}$, which is expressed by:

\begin{equation}\label{eq:gate1}
\mathcal{G}_{(X,Y)} = \sum_{x \in X}\sum_{m \in \mathcal{M}_{\text{enc}}} g_m (x) + \sum_{y \in Y}\sum_{m \in \mathcal{M}_{\text{dec}}} g_m (y)
\end{equation} 

where $\mathcal{M}_{\text{enc}}$ and $\mathcal{M}_{\text{dec}}$ are respectively the sets of encoders and decoders layers, and $g_m(.)=1$ when LS expert is selected in the layer $m$, or $g_m(.)=0$ otherwise. The average of this gate usage, representing the amount of language-specific experts used for the model in the batch, is constrained to a budget $b$. So the final gate budget loss is expressed by: 

\begin{equation}\label{eq:gateloss}
    \mathcal{L}_{\text{g}} = \left\lvert \frac{\sum_{(X,Y) \in \mathcal{B}} \mathcal{G}_{(X,Y)}}{\sum_{(X,Y) \in \mathcal{B}} (|X||\mathcal{M}_{\text{enc}}| + |Y||\mathcal{M}_{\text{dec}}|)} - b \right\rvert
\end{equation}

For regularization, also it is used a skip gate probability ($s$), that randomly choose a proportion $s$ of the gates to be closed (use only shared part) during training. 

\subsection{Knowledge Distillation}

For Knowledge Distillation (KD), following recent research~\citep{js_vs_kd_acl23,js_vs_kd_bb2023}, we employ Jensen–Shannon divergence (JS), whose loss is detailed in Eq~\ref{eq:kd}:
\begin{equation}\label{eq:kd}
    \mathcal{L}_{\text{KD}} = \frac12 \mathbb{E}_{\mathbf Y \sim p} \left[ \log \tfrac{p(\mathbf Y)}{m(\mathbf Y)} \right] + \frac12 \mathbb{E}_{\mathbf Y' \sim q_\theta} \left[ \log \tfrac{q_\theta (\mathbf Y')}{m(\mathbf Y')} \right]
\end{equation}
where $p$ is the teacher distribution, $q_\theta$ is the student distribution, $\mathbf Y$ and $\mathbf Y'$ are sampled from the teacher's and student's distributions and compared with their average $m(\cdot)=\frac12 p(\cdot) + \frac12q_\theta(\cdot)$. 


\subsection{Final Learning Objective}

The final learning objective the leverages the dataset labels using cross-entropy loss $\mathcal{L}_{\text{CE}}$, but also enforces the use of a specific budget via gate budget loss $\mathcal{L}_{\text{g}}$ and mirrors the behavior of the teacher with the knowledge distillation loss $\mathcal{L}_{\text{KD}}$.Thus, CLSR modules parameters are learned to minimize final loss expressed as: 
\begin{equation}
\mathcal{L} = \mathcal{L}_{\text{CE}} + \mathcal{L}_{\text{g}} + \beta\mathcal{L}_{\text{KD}}
\end{equation}
where $\beta$ is a constant defined based on the quality of the teacher, but can also be scheduled or learned (with the add of new constraints for its magnitude).

            \chapter{Experiments and Results on DistilWhisper}\label{chap:results}

In the former chapter we presented the \textit{DistilWhisper} approach. In this chapter we present how we validate our architecture and the method as a whole, showing that our approach is able to outperform both classical fine-tuning and adapters on \texttt{whisper-small}, providing better generalization through light-weight ASR fine-tuning and Knowledge Distillation of the teacher model. Code and models produced in this studied will soon be made available on Hugging Face and Github.

\section{Experimental Setup}\label{sec:setup}

In this section we overview our validation setup, that includes choosing the data we use for training and evaluating models, as well as which languages and baselines to consider. We also discuss some code implementation details.

\subsection{Datasets} 

In order to validate the proposed architecture, we make use of a sample of two widely used massively-multilingual datasets: CommonVoice~13.0 and FLEURS. More details about these datasets are presented on Section \ref{sec:dataset-background}.

In our experiments, we applied downsampling to both the train and validation sets of CV-13, ensuring an equal allocation of training data for each selected language in each experiment. For our primary experiment, we employed 10,000 utterances for training (approximately 14 hours of audio data) and 1,000 for validation. Additionally, we explored variations in dataset size, using downsampled sets of 3,000 and 28,000 utterances in scalability experiments. The selection of data for downsampling was guided by the number of up-votes received by annotators. Notably, we did not apply downsampling to the test set.

For most part of our experiments, FLEURS serves as an invaluable resource for conducting out-of-domain evaluations. It offers a favorable degree of language overlap with the CommonVoice 13.0 dataset (CV-13), making it a suitable choice for comparative analysis. Notably, FLEURS provides an effective out-of-domain setting in the context of ASR evaluation. For instance, while the average number of tokens per sample in CV-13 is 36, FLEURS exhibits a substantially higher average of 97 tokens per sample.

\subsection{Language Selection} 
In this work we focus on bridging the performance gap for a subset of under-performing languages of the \texttt{whisper-small} model through light-weight ASR fine-tuning and Knowledge Distillation of the \texttt{whisper-large-v2} model, as proposed in chapter \ref{chap:distilwhisper}. For validating our method, we consider all Whisper languages with a WER gap of more than 11 between large and small models on CV-13.

For our validation experiments we then narrow this list considering: 1) minimum amount of 10k utterances; 2) an overlap with the FLEURS dataset for out-of-domain evaluation. For scalability experiments we loose the first requirement so we can include more diverse set of languages, considering a minimum amount of 3k utterances. We also experiment with the languages in a setting with 28k utterances. 

\begin{table}[caption={Languages used in the experiments for validation of \textit{DistilWhisper} grouped by resourcefulness.},label=tab:resourcefulness]
\begin{tabular}{@{}c|c|ccc@{}}
\toprule
\multirow{2}{*}{Resourcefulness} & \multirow{2}{*}{ASR Train data (h)} & \multicolumn{3}{c}{Languages per setting}    \\
                                 &                                        & 3k                 & 10k            & 28k    \\ \midrule
High-resource                    & {[}1000, 5000)                         & ca, fi, id, pl             & ca, pl         & ca     \\
Mid-to-high-resource             & {[}500, 1000)                          & uk, vi             & uk             & -      \\
Low-to-mid-resource              & {[}100, 500)                           & cs, hu, ro, th, ta & cs, hu, th, ta & ta, th \\
Low-resource                     & {[}10, 100)                            & bg, hi, sk, sl     & -              & -      \\
Extremely Low-Resource           & (0, 10)                                & gl                 & gl             & -      \\ \bottomrule
\end{tabular}
\end{table}


The final list of languages is: Bulgarian~(bg), Catalan~(ca), Czech~(cs), Finnish~(fi), Galician~(gl), Hindi~(hi), Hungarian~(hu), Indonesian~(id), Polish~(pl), Romanian~(ro), Slovak~(sk), Slovenian~(sl), Tamil~(ta), Thai~(th), Ukranian~(uk), and Vietnamease~(vi).\footnote{Although Arabic would also qualify considering our criteria, we find that the dialect from FLEURS differs from the ones present on CV-13.} These languages belong to 7 distinct language sub-families and exhibit significant variation in terms of their representation within the Whisper training data. This variation extends from a substantial 4,300 hours for certain languages, such as Polish (pl), to a mere 9 hours for languages like Galician (gl). For a detailed overview of these languages and their distribution across the three dataset sizes (3k, 10k, 28k), categorized by their resourcefulness (following the classification proposed on Section \ref{sec:resourcefulnessclass}), please refer to Table \ref{tab:resourcefulness}. Additionally, Table \ref{tab:subfamilies} organizes these languages into groups based on their respective sub-families. 

\begin{table}[caption={Languages used in the experiments for validation of \textit{DistilWhisper} grouped by language sub-families.},label=tab:subfamilies]
\begin{tabular}{@{}cccc@{}}
\toprule
\multirow{2}{*}{Sub-families}                                 & \multicolumn{3}{c}{Languages per setting}                                                                                         \\ \cmidrule(l){2-4} 
                                                             & 3k                                                                                 & 10k                     & 28k                \\ \midrule
\multicolumn{1}{c|}{\multirow{2}{*}{Slavic (Indo-European)}} & \multirow{2}{*}{\begin{tabular}[c]{@{}c@{}}bg, cs, pl, \\ sk, sl, uk\end{tabular}} & \multirow{2}{*}{cs, pl} & \multirow{2}{*}{-} \\
\multicolumn{1}{c|}{}                                        &                                                                                    &                         &                    \\
\multicolumn{1}{c|}{Romance (Indo-European)}                 & ca, gl, ro                                                                         & ca, gl                  & ca                 \\
\multicolumn{1}{c|}{Finno-Ugrian (Uralic)}                   & fi, hu                                                                             & hu                      & -                  \\
\multicolumn{1}{c|}{Austroasiatic}                           & id, vi                                                                             & -                       & -                  \\
\multicolumn{1}{c|}{Dravidian}                               & ta                                                                                 & ta                      & ta                 \\
\multicolumn{1}{c|}{Tai (Kra–Dai)}                           & th                                                                                 & th                      & th                 \\
\multicolumn{1}{c|}{Indo-Iranian (Indo-European)}            & hi                                                                                 & -                       & -                  \\ \bottomrule
\end{tabular}

\end{table}


\subsection{Models and Baselines}

In our evaluation, we assess our approach in comparison to several baseline models. These include the \texttt{whisper-small} model, serving as our pre-trained student and starting point, and the \texttt{whisper-large-v2} model, acting as the teacher model, and ultimately, as the target goal. Additionally, we explore two fine-tuning (FT) approaches for the student model: standard fine-tuning, where all model weights are updated, and LoRA adaptation, which focuses on refining the feed-forward layer. Moreover, we delve into the effects of the Conditional Language-Specific Routing (CLSR) layer independently, without knowledge distillation (KD), referred to as CLSR-FT. This allows us to isolate the influence of KD from the impact of the CLSR layer on the model's overall robustness.

\subsection{Implementation details} 

We conducted our experiments using the Transformers library~\citep{wolf2020transformers} and leveraged the pre-trained weights of both \texttt{whisper-small} and \texttt{whisper-large-v2} models, sourced from HuggingFace\footnote{\url{https://huggingface.co/openai/}} \footnote{\url{https://huggingface.co/collections/openai/whisper-release-6501bba2cf999715fd953013}}. Unless where stated different, our training protocol consisted of ten epochs, utilizing a learning rate of $10^{-4}$ with linear decay, a one-epoch warm-up phase, a batch size of 16, and a label smoothing factor of 0.1.

For LoRA adaptation, we tested two scenarios: 1) We first adopted the hyperparameters proposed by~\citet{wang23ga_interspeech}, notably $r = 32$, which is the most commonly used for this type of adapters; 2) We increase the hidden dimension of the adapters to $r = 64$, so the size of the adapters are comparable to the Language-specific modules on \textit{DistilWhisper}. 

In the case of training the CLSR, we set the gate budget ($b$) to 0.5 and the skip-gate probability ($s$) to 0.2. For knowledge distillation (KD), we employed the Jensen–Shannon divergence (JS) with a temperature ($\tau$) of 1, unless when stated in contrary. This was weighted such that the learning objective ($\mathcal{L}$) consisted of the cross-entropy loss ($\mathcal{L}_{\text{CE}}$), the gate loss ($\mathcal{L}_{\text{g}}$), and twice the KD loss ($2\mathcal{L}_{\text{KD}}$): $\mathcal{L} = \mathcal{L}_{\text{CE}} +\mathcal{L}_{\text{g}} + 2\mathcal{L}_{\text{KD}}$.

We reported the normalized Word Error Rate (WER) using the Whisper normalization method, with a slight modification to prevent the splitting of numbers and Latin-scripted text into individual characters in languages that do not employ space delimitation (e.g., Thai). Further details, including the modified normalization method, implementation scripts, and model weights, will soon be made available on GitHub and HuggingFace.

Throughout our experiments, we selected the best-performing model based on its WER performance on the downsampled CV-13 validation set.




\section{\textit{DistilWhisper} versus other adaptation approaches} \label{sec:distilwhispervsothers}

Table~\ref{tab:results1} presents the results for our first experiment. The top portion presents \texttt{whisper-large-v2} (upper bound) and \texttt{whisper-small} (lower bound) pre-trained scores, which should not be directly compared to the other adaptation techniques~(middle and bottom), as these models were not trained on CV-13~(full out-of-domain setting). The middle portion presents standard fine-tuning~(FT) and LoRA adaptation at the feed-forward layers~(LoRA-FT). Our results are presented in the bottom: CLSR-FT corresponds to the setting without $\mathcal{L}_{\text{KD}}$, while \textit{DistilWhisper} is the complete setting in which both CLSR and KD losses are leveraged.

\begin{table}[caption={WER~($\downarrow$) for the 10k setting with dataset averages~(avg) for baselines~(top), adaptation approaches~(middle), and our method~(bottom) for in-domain~(CV-13, FT only) and out-of-domain~(FLEURS, all) test sets.Best results for \texttt{whisper-small} in \textbf{bold}.},label=tab:results1]
\resizebox{\textwidth}{!}{
\begin{tabular}{lcccccccccc}
\hline
\multicolumn{11}{c}{\textbf{Common voice 13.0 (in-domain for FT only)}}                                                                                                                                     \\ \hline
\multicolumn{1}{c}{\textbf{Model}}     & \textbf{\# params} & \textbf{avg}                                              & \textbf{ca}                          & \textbf{th}                          & \textbf{ta}                          & \textbf{hu}                          & \textbf{cs}                          & \textbf{pl}                          & \textbf{gl}                          & \textbf{uk}                          \\ \hline
whisper large-v2             & 1.5 B         & \multicolumn{1}{c|}{{\color[HTML]{9B9B9B} 14.9}} & {\color[HTML]{9B9B9B} 16.9} & {\color[HTML]{9B9B9B} 9.3}  & {\color[HTML]{9B9B9B} 17.3} & {\color[HTML]{9B9B9B} 18.6} & {\color[HTML]{9B9B9B} 14.5} & {\color[HTML]{9B9B9B} 8.1}  & {\color[HTML]{9B9B9B} 19.0} & {\color[HTML]{9B9B9B} 15.6} \\
whisper-small                & 244 M         & \multicolumn{1}{c|}{{\color[HTML]{9B9B9B} 31.4}} & {\color[HTML]{9B9B9B} 30.1} & {\color[HTML]{9B9B9B} 20.3} & {\color[HTML]{9B9B9B} 30.1} & {\color[HTML]{9B9B9B} 45.5} & {\color[HTML]{9B9B9B} 38.6} & {\color[HTML]{9B9B9B} 18.8} & {\color[HTML]{9B9B9B} 35.7} & {\color[HTML]{9B9B9B} 32.3} \\ \hline
$\quad$+FT             & 244 M         & \multicolumn{1}{c|}{22.0}                        & 19.0                        & 10.9                        & 17.3                        & 30.4                        & 29.2                        & 21.4                        & 19.3                        & 28.8                        \\
$\quad$+LoRA-FT (r=32) & 256 M         & \multicolumn{1}{c|}{18.6}                        & 15.7                        & 9.2                         & 15.3                        & 30.5                        & 25.0                        & \textbf{15.4}               & 12.8                        & 24.8                        \\
$\quad$+LoRA-FT (r=64) & 267 M         & \multicolumn{1}{c|}{18.6}                        & 15.5                        & 9.2                         & 15.5                        & 30.6                        & 25.2                        & \textbf{15.4}               & 13.0                        & 24.6                        \\ \hline
$\quad$+CLSR-FT        & 269 M         & \multicolumn{1}{c|}{16.4}                        & 13.9                        & 7.4                         & 13.6                        & 24.9                        & 20.9                        & 16.0                        & \textbf{11.2}               & 23.5                        \\
DistilWhisper                & 269 M         & \multicolumn{1}{c|}{\textbf{16.1}}               & \textbf{13.8}               & \textbf{7.2}                & \textbf{12.5}               & \textbf{24.1}               & \textbf{19.9}               & 16.1                        & 11.6                        & \textbf{23.2}               \\ \hline
\multicolumn{11}{c}{\textbf{FLEURS (out-of-domain)}}                                                                                                                                                                                                                                                                                                     \\ \hline
\multicolumn{1}{c}{\textbf{Model}}    & \textbf{\# params} & \textbf{avg}                                              & \textbf{ca}                          & \textbf{th}                          & \textbf{ta}                          & \textbf{hu}                          & \textbf{cs}                          & \textbf{pl}                          & \textbf{gl}                          & \textbf{uk}                          \\ \hline
whisper large-v2             & 1.5 B         & \multicolumn{1}{c|}{12.6}                        & 5.6                         & 12.6                        & 19.3                        & 17.9                        & 14.4                        & 5.9                         & 16.8                        & 8.3                         \\
whisper-small                & 244 M         & \multicolumn{1}{c|}{29.2}                        & 14.6                        & 22.7                        & 36.2                        & 42.9                        & 40.3                        & \textbf{18.2}                        & 33.5                        & 24.8                        \\ \hline
$\quad$+FT             & 244 M         & \multicolumn{1}{c|}{30.8}                        & 19.1                        & 28.2                        & 31.6                        & 51.3                        & 38.9                        & 26.1                        & 23.2                        & 27.9                        \\
$\quad$+LoRA-FT (r=32) & 256 M         & \multicolumn{1}{c|}{23.6}                        & 15.5                        & 17.6                        & 25.5                        & 38.5                        & 33.4                        & 18.5               & 17.7                        & 22.3                        \\
$\quad$+LoRA-FT (r=64) & 267 M         & \multicolumn{1}{c|}{23.6}                        & 15.7                        & 17.6                        & 25.7                        & 38.2                        & 33.9                        & 18.5               & 17.3                        & \textbf{22.1}               \\ \hline
$\quad$+CLSR-FT        & 269 M         & \multicolumn{1}{c|}{23.6}                        & 15.5                        & 15.7                        & 23.2                        & 37.6                        & 31.2                        & 22.9                        & 16.9                        & 25.9                        \\
DistilWhisper                & 269 M         & \multicolumn{1}{c|}{\textbf{22.8}}               & \textbf{15.4}               & \textbf{15.1}               & \textbf{21.6}               & \textbf{37.2}               & \textbf{29.8}               & 21.4                        & \textbf{16.7}               & 25.1                       
\end{tabular}
}
\end{table}

For \texttt{whisper-small}, we observe that both the standard fine-tuning method (FT) and the LoRA Adapters~(LoRA-FT) approaches~(middle portion of Table~\ref{tab:results1}) demonstrate the capacity to enhance performance on the in-domain test set (CV-13). However, as anticipated, employing FT leads to a decline in performance on the out-of-domain test set, with an average increase of 1.6. This is likely attributed to catastrophic forgetting, resulting in a tendency to overly specialize in the specific domain.
In contrast, LoRA-FT represents a more lightweight adaptation technique that preserves the pre-trained representation. Remarkably, it exhibits improvements in performance on both the in-domain (average decrease of 12.8) and out-of-domain (average decrease of 5.6) test sets when compared to \texttt{whisper-small}. Notably, experimenting with a larger hidden dimension ($r$) for the LoRA adapters did not yield any perceptible improvement on the average.


Our approach, \textit{DistilWhisper}, yields notable enhancements in performance. When compared to \texttt{whisper-small}, it achieves a substantial improvement on in-domain data, with an average decrease of 15.3. This improvement is also evident when compared to LoRA-FT, where an average decrease of 2.2 is observed. Additionally, \textit{DistilWhisper} exhibits superior adaptability in out-of-domain scenarios when contrasted with the original \texttt{whisper-small}, resulting in an average increase of 6.4. Furthermore, it demonstrates more effective out-of-domain adaptation capabilities in comparison to LoRA-FT, with an average increase of 0.8. We observe that both versions of our approach, with and without KD, successfully outperform all other adaptation approaches~(FT, LoRA-FT) for in-domain and out-of-domain in all languages but two (pl and uk) (bottom portion of Table~\ref{tab:results1}). These findings highlights the robustness of our approach, showcasing that the proposed architecture with the addition of CLSR layers on Whisper provides a strong solution. Notably, all of these improvements are achieved with a mere 25 million parameter overhead during inference (10 \% of the original model size). 



\section{Impact of knowledge distillation} \label{sec:effect-kd}

In this analysis, we compare the two versions of our approach: one entails optimizing a lightweight CLSR-based architecture without Knowledge Distillation (CLSR-FT), while the other incorporates Knowledge Distillation loss (\textit{DistilWhisper}). Across the examined languages, we observe some interesting trends.

Firstly, when considering in-domain performance, as shown in Table~\ref{tab:results1}, the \textit{DistilWhisper} model exhibits a slightly increase in average performance of 0.3 on the WER. The performance is superior in all languages but Polish and Galician. However, when it comes to out-of-domain scenarios, \textit{DistilWhisper} consistently outperforms CLSR-FT across all languages, resulting in an average improvement of 0.8 on the WER. This observation confirms our initial hypothesis that the inclusion of Knowledge Distillation leverages the robustness imparted by the teacher model, preventing over-specialization in the CV-13 domain.

Collectively, these results underscore the effectiveness of our proposed architecture. Notably, we managed to bridge the out-of-domain performance gap between \texttt{large-v2} and \texttt{small} by a substantial 39\%, reducing it from 16.6 to 10.2 (average decrease of 6.5). All of this was achieved with only a modest 10\% parameter overhead during inference (25 million parameters).


\section{\textit{DistilWhisper} Scalability} \label{sec:scalability}
%

In the previous sections we showed that our architecture improves scores for both in-domain and out-of-domain datasets, compared to other adaptation approaches. In this section we investigate the effectiveness of our method with respect to the amount of data available for training. For this, we select a subset of languages for which we find more training data available on CV-13~(ca, th, ta). Table~\ref{tab:results2} presents results for our approach in lower-resource training settings~(3k utterances; approx. 4 hours), and higher-resource settings~(28k utterances; approx. 40 hours). 10k results as well as the results for \texttt{whisper-large-v2} and \texttt{whisper-small} are repeated from Table~\ref{tab:results1}.

\begin{table}[caption={WER~($\downarrow$) for different training data sizes~(3k, 10k, and 28k utterances) for both in-domain~(CV-13) and out-of-domain~(FLEURS) test sets. Best results in \textbf{bold}.},label=tab:results2]
\resizebox{\textwidth}{!}{
\begin{tabular}{@{}lccc|ccc|ccc@{}}
\toprule
\multicolumn{1}{c}{}         & \textbf{}      & \textbf{}       & \textbf{}      & \multicolumn{3}{c|}{\textbf{FLEURS}}          & \multicolumn{3}{c}{\textbf{CV-13}}           \\
\multicolumn{1}{c}{}         & \textbf{Train} & \textbf{FLEURS} & \textbf{CV-13} & \multicolumn{3}{c|}{\textbf{(out-of-domain)}} & \multicolumn{3}{c}{\textbf{(in-domain)}}     \\
\multicolumn{1}{c}{}         & \textbf{size}  & \textbf{avg}    & \textbf{avg}   & \textbf{ca}   & \textbf{ta}   & \textbf{th}   & \textbf{ca}   & \textbf{ta}   & \textbf{th}  \\ \midrule
whisper large-v2             & -              & 12.5            & 14.5           & 5.6           & 19.3          & 12.6          & 16.9          & 17.3          & 9.3          \\
whisper-small                & -              & 24.5            & 26.8           & 14.6          & 36.2          & 22.7          & 30.1          & 30.1          & 20.3         \\ \midrule
$\quad$+LoRA-FT (r=64) & 3k             & 22.7            & 17.0           & 17.7          & 28.6          & 21.7          & 19.4          & 19.0          & 12.5         \\
$\quad$+CLSR-FT        & 3k             & 20.4            & 15.2           & 17.8          & \textbf{25.4} & 17.9          & 19.2          & 16.7          & 9.7          \\
DistilWhisper                & 3k             & \textbf{20.2}   & \textbf{14.8}  & \textbf{17.2} & 25.7          & \textbf{17.6} & \textbf{18.9} & \textbf{15.9} & \textbf{9.6} \\ \midrule
$\quad$+LoRA-FT (r=64) & 10k            & 19.7            & 13.4           & 15.7          & 25.7          & 17.6          & 15.5          & 15.5          & 9.2          \\
$\quad$+CLSR-FT        & 10k            & 18.1            & 11.6           & 15.5          & 23.2          & 15.7          & 13.9          & 13.6          & 7.4          \\
DistilWhisper                & 10k            & \textbf{17.4}   & \textbf{11.2}  & \textbf{15.4} & \textbf{21.6} & \textbf{15.1} & \textbf{13.8} & \textbf{12.5} & \textbf{7.2} \\ \midrule
$\quad$+LoRA-FT (r=64) & 28k            & 17.2            & 11.1           & 13.6          & 23.0          & 15.1          & 12.5          & 13.5          & 7.3          \\
$\quad$+CLSR-FT        & 28k            & 15.6            & 9.7            & 13.5          & 19.6          & \textbf{13.8} & 11.5          & 11.3          & 6.2          \\
DistilWhisper                & 28k            & \textbf{15.4}   & \textbf{9.3}   & \textbf{13.1} & \textbf{19.2} & 14.0          & \textbf{11.3} & \textbf{10.9} & \textbf{5.7} \\ \bottomrule
\end{tabular}
}
\end{table}


We observe that, as expected, increasing the amount of trainable examples leads to superior ASR performance for both approaches, with the leveraging of KD~(\textit{DistilWhisper}) being consistently superior to CLSR-FT and getting closer to close the out-of domain performance gap. For the 28k setup, we are able to reduce the out-of-domain WER gap between \texttt{whisper-large-v2} and \texttt{whisper-small} by 75.8\%, from 12.0 to 2.9. 

Furthermore, our approach demonstrates commendable robustness in relation to the quantity of trainable examples. Even with as few as 3,000 utterances (equivalent to 4 hours of training data), we are able to reduce the WER performance gap by 35.8\% in out-of-domain data. This suggests that our method holds promise in enhancing ASR performance for low-resource languages, where training data availability is limited.

Across all three settings, our approaches consistently outperform LoRA Adapters by a significant margin. Additionally, it is worth noting that, in nearly all cases within these settings, the inclusion of knowledge distillation proved more beneficial than fine-tuning alone, reinforcing the findings discussed in Section \ref{sec:effect-kd}.





\section{Gate Activation Analysis}

To better understand how the model uses routing mechanism, we analyze gate activation statistics on the experiment discussed on Section \ref{sec:scalability} for both CLSR-FT and \textit{DistilWhisper}. This results are presented on Figure~\ref{fig:gates}.

\begin{figure}[caption={Ratio of LS layers chosen by the models~(CLSR-FT and \textit{DistilWhisper}) depending on (1) amount of training data; 
    (2) in (CV-13) or out-of-domain (FLEURS); (3) language. 
    },label=fig:gates]
    \centering
    \includegraphics[width=\linewidth]{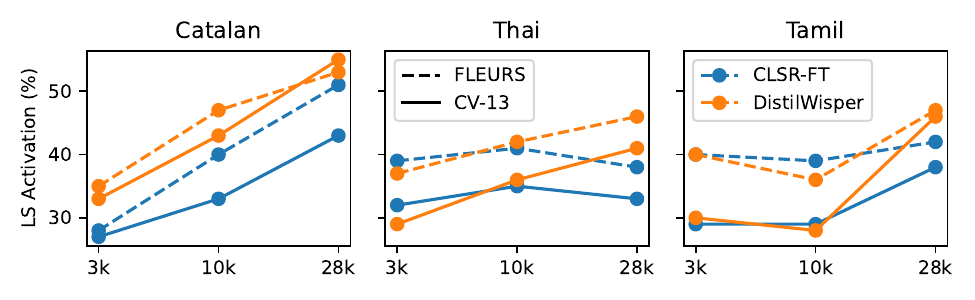}
\end{figure}

Firstly, we observe a tendency for the models to rely more heavily on the newly introduced Language-Specific modules in out-of-domain scenarios. This could be attributed to the greater complexity and larger sentence sizes prevalent in the FLEURS dataset. 

Also, as expected, enlarging the training dataset consistently results in more reliable Language-Specific modules, leading to increased utilization of these modules. The only exception for this is Thai at the 28k setup with CLSR-FT, and this might be due to dataset quality and requires further investigation

The comparison of the three languages reveals that Catalan displays a notably higher reliance on Language-Specific routes. This characteristic might be linked to the superior data quality available for Catalan in CV-13, where a substantial number of contributors have contributed to the dataset. Also, the distilled version uses more LS modules, probably because the teacher \texttt{whisper-large-v2} is a really good model for this language.

Now for languages with a weaker teacher~(Thai, Tamil) we observe that the model may receive contradictory signals at lower-resource settings~(3k, 10k), leading to less Language-Specific routing usage with Knowledge Distilation. However, in the higher resource setting~(28k), KD usage leads systematically to more reliable Language-Specific module and therefore higher LS routing.

Finally, we observe a common trend across the three languages - models tend to employ more Language-Specific routes when learning with Knowledge Distillation (\textit{DistilWhisper} vs. CLSR-FT). This suggests that KD imparts valuable information and enhances the out-of-domain generalization capabilities of the learned Language-Specific representation.

\begin{landscape}
\begin{table}[caption={WER~($\downarrow$) for the 3k setting with dataset averages~(avg) for baselines~(top), and our method~(bottom) for in-domain~(CV-13, FT only - higher portion) and out-of-domain~(FLEURS, all - middle portion) test sets. On the lower portion, the same results are grouped by resourcefulness. Best results for \texttt{whisper-small} in \textbf{bold}.},label=tab:results3]
\begin{tabular}{@{}lcccccccccccccccc@{}}
\toprule
\multicolumn{17}{c}{\textbf{Common voice 13.0 (in-domain for FT only)}}                                                                                                                                                                                                                                                                                                                                                                                                                                                                                                 \\ \midrule
\multicolumn{1}{c}{\textbf{Model}} & \multicolumn{1}{c|}{\textbf{avg}}                & \textbf{bg}                  & \textbf{ca}                 & \textbf{cs}                 & \textbf{fi}                  & \textbf{gl}                 & \textbf{hi}                  & \textbf{hu}                 & \textbf{id}                  & \textbf{pl}                 & \textbf{ro}                  & \textbf{sk}                           & \textbf{sl}                  & \textbf{ta}                 & \textbf{th}                 & \textbf{uk}                 \\ \midrule
whisper large-v2                   & \multicolumn{1}{c|}{{\color[HTML]{656565} 17.0}} & {\color[HTML]{656565} 19.9}  & {\color[HTML]{656565} 16.9} & {\color[HTML]{656565} 14.5} & {\color[HTML]{656565} 14.4}  & {\color[HTML]{656565} 19.0} & {\color[HTML]{656565} 24.6}  & {\color[HTML]{656565} 18.6} & {\color[HTML]{656565} 8.5}   & {\color[HTML]{656565} 8.1}  & {\color[HTML]{656565} 15.8}  & {\color[HTML]{656565} 31.9}           & {\color[HTML]{656565} 20.6}  & {\color[HTML]{656565} 17.3} & {\color[HTML]{656565} 9.3}  & {\color[HTML]{656565} 15.6} \\
whisper-small                      & \multicolumn{1}{c|}{{\color[HTML]{656565} 34.2}} & {\color[HTML]{656565} 44.8}  & {\color[HTML]{656565} 30.1} & {\color[HTML]{656565} 38.6} & {\color[HTML]{656565} 30.5}  & {\color[HTML]{656565} 35.7} & {\color[HTML]{656565} 43.6}  & {\color[HTML]{656565} 45.5} & {\color[HTML]{656565} 22.5}  & {\color[HTML]{656565} 18.8} & {\color[HTML]{656565} 33.2}  & {\color[HTML]{656565} 42.0}           & {\color[HTML]{656565} 45.5}  & {\color[HTML]{656565} 30.1} & {\color[HTML]{656565} 20.3} & {\color[HTML]{656565} 32.3} \\ \midrule
$\quad$+CLSR-FT              & \multicolumn{1}{c|}{22.9}                        & 26.1 & 19.2                        & \textbf{25.7}               & 25.1 & 15.3                        & 18.8 & 31.6                        & 19.2 & \textbf{18.3}               & 23.4 & \textbf{36.6} & 28.6 & 16.7                        & 9.7                         & \textbf{29.5}               \\
DistilWhisper                      & \multicolumn{1}{c|}{\textbf{22.6}}               & \textbf{25.9}                & \textbf{18.9}               & 26.2                        & \textbf{24.8}                & \textbf{14.7}               & \textbf{18.3}                & \textbf{31.0}               & \textbf{18.6}                & 18.6                        & \textbf{21.5}                & 36.8                                  & \textbf{27.7}                & \textbf{15.9}               & \textbf{9.6}                & 30.0                        \\ \midrule
\multicolumn{17}{c}{\textbf{FLEURS (out-of-domain)}}                                                                                                                                                                                                                                                                                                                                                                                                                                                                                                                    \\ \midrule
\multicolumn{1}{c}{\textbf{Model}} & \multicolumn{1}{c|}{\textbf{avg}}                & \textbf{bg}                  & \textbf{ca}                 & \textbf{cs}                 & \textbf{fi}                  & \textbf{gl}                 & \textbf{hi}                  & \textbf{hu}                 & \textbf{id}                  & \textbf{pl}                 & \textbf{ro}                  & \textbf{sk}                           & \textbf{sl}                  & \textbf{ta}                 & \textbf{th}                 & \textbf{uk}                 \\ \midrule
whisper large-v2                   & \multicolumn{1}{c|}{13.7}                        & 14.6                         & 5.6                         & 14.4                        & 9.7                          & 16.8                        & 23.8                         & 17.9                        & 7.1                          & 5.9                         & 14.4                         & 11.7                                  & 23.1                         & 19.3                        & 12.6                        & 8.3                         \\
whisper-small                      & \multicolumn{1}{c|}{32.8}                        & 39.9                         & 14.6                        & 40.3                        & 26.8                         & 33.5                        & 47.9                         & 42.9                        & \textbf{18.6}                & \textbf{18.2}               & 34.6                         & 35.8                                  & 54.5                         & 36.2                        & 22.7                        & 24.8                        \\ \midrule
$\quad$+CLSR-FT              & \multicolumn{1}{c|}{\textbf{29.2}}               & 43.8 & 17.8                        & 35.4                        & 33.7 & 19.8                        & 22.8 & \textbf{40.1}               & 19.0 & 21.9                        & 33.4 & 35.3          & 50.8 & \textbf{25.4}               & 17.9                        & \textbf{21.6}               \\
DistilWhisper                      & \multicolumn{1}{c|}{\textbf{29.2}}               & \textbf{42.8}                & \textbf{17.2}               & \textbf{35.6}               & \textbf{32.0}                & \textbf{18.7}               & \textbf{21.8}                & 41.1                        & 19.1                         & 21.9                        & \textbf{33.1}                & \textbf{35.2}                         & \textbf{50.5}                & 25.7                        & \textbf{17.6}               & 25.9                        \\ \bottomrule
\end{tabular}

\bigskip

\resizebox{\columnwidth}{!}{
\begin{tabular}{@{}lccccc|ccccc@{}}
\toprule
\textbf{}             & \multicolumn{5}{c|}{\textbf{FLEURS (out-of-domain)}}                                                                                                                                                    & \multicolumn{5}{c}{\textbf{CV-13 (in-domain)}}                                                                                                                                                         \\ \midrule
\textbf{}             & \multicolumn{1}{l}{\textbf{High}} & \multicolumn{1}{l}{\textbf{Mid-to-high}} & \multicolumn{1}{l}{\textbf{Low-to-mid}} & \multicolumn{1}{l}{\textbf{Low}} & \multicolumn{1}{l|}{\textbf{Extremely Low}} & \multicolumn{1}{l}{\textbf{High}} & \multicolumn{1}{l}{\textbf{Mid-to-high}} & \multicolumn{1}{l}{\textbf{Low-to-mid}} & \multicolumn{1}{l}{\textbf{Low}} & \multicolumn{1}{l}{\textbf{Extremely Low}} \\ \midrule
whisper large-v2      & 7.1                               & 8.3                                      & 15.7                                    & 18.3                             & 16.8                                        & 12.0                              & 15.6                                     & 15.1                                    & 24.3                             & 19.0                                       \\
whisper-small         & \textbf{19.6}                     & 24.8                                     & 35.3                                    & 44.5                             & 33.5                                        & 25.5                              & 32.3                                     & 33.5                                    & 44.0                             & 35.7                                       \\ \midrule
$\quad$+CLSR-FT & 23.1                              & \textbf{21.6}                            & \textbf{30.4}                           & 38.2                             & 19.8                                        & 20.4                              & \textbf{29.5}                            & 21.4                                    & 27.5                             & 15.3                                       \\
DistilWhisper         & 22.5                              & 25.9                                     & \textbf{30.6}                           & \textbf{37.6}                    & \textbf{18.7}                               & \textbf{20.2}                     & 30.0                                     & \textbf{20.8}                           & \textbf{27.2}                    & \textbf{14.7}                              \\ \bottomrule
\end{tabular}
}
\end{table}

\end{landscape}

\section{Considerations on the Resourcefulness}

Our observations so far indicate that both versions of our approach, with and without knowledge distillation (KD), demonstrate consistent outperformance over all other adaptation methods (FT and LoRA-FT). This improvement holds true for both in-domain and out-of-domain scenarios across all languages, with only two exceptions on the 10k setting (Polish and Ukrainian), as indicated in the lower portion of Table~\ref{tab:results1}. The challenges encountered in these two languages can be attributed to their higher resource status, with Polish being a high-resource language and Ukrainian categorized as mid-to-high resource, as detailed in Table~\ref{tab:resourcefulness}.

In order to deepen this analysis, we conducted experiments across a broader range of languages, widening to those with a minimum of 3,000 utterances available for training. The outcomes of these experiments are presented in Table~\ref{tab:results3}, where we have also aggregated the results into resourcefulness clusters (in the lower portion) based on the classification provided in Table~\ref{tab:resourcefulness}. 

Examining the results, we observed that more substantial out-of-domain improvements are seem in languages with lower resource availability (Low-to-mid, Low and Extremely low-resource clusters). This aligns with the initial motivation behind our work, which aimed to address the \textit{curse of multilinguality}. We expect that lower resource languages experience a more significant impact from this phenomenon during the pre-training of \texttt{whisper-small}. Consequently, they significantly benefit more from the integration of language-specific modules in the feature domain.

In contrast, for languages with higher resource availability, further enhancements may be necessary, such as adjustments to attention weights (corresponding to the time domain). This is due to the fact that the original model already performs reasonably well. Additionally, achieving better out-of-domain performance may require a larger training dataset. This is exemplified by the case of Catalan presented in Table~\ref{tab:results2}. In this case, CLSR modules yielded superior performance than original \texttt{whisper-small} only in the case trained with 28,000 utterances, losing to its starting point for 3,000 and 10,000 training utterances.

\section{Effect of temperature and distillation loss}

In this set of experiments, our goal is to examine the impact of the chosen distillation optimization on the results. We start by exploring the effect of temperature. Temperature plays a crucial role in determining the learning behavior of the model. A lower temperature, like 1, tends to make the learning focus primarily on replicating the first option from the teacher's logits for each token. Conversely, a higher temperature, such as 3 or 4, encourages the learning to take into account the other options, thereby mitigating the cost from incorrect predictions. However, this approach may lead to over-smoothing of the distribution and a reduced ability to effectively rank similar logits.

\begin{table}[caption={WER~($\downarrow$) for the 10k setting with dataset averages~(avg) for JS loss with temperatures 1 and 3, for in-domain~(CV-13, FT only - higher portion) and out-of-domain~(FLEURS, all - lower portion) test sets. Best results in \textbf{bold}.},label=tab:results4]
\begin{tabular}{@{}lccccccccc@{}}
\toprule
\multicolumn{10}{c}{\textbf{Common voice 13.0 (in-domain)}}                                                                                                                          \\ \midrule
                & \multicolumn{1}{c|}{\textbf{avg}}  & \textbf{ca}   & \textbf{th}   & \textbf{ta}   & \textbf{hu}   & \textbf{cs}   & \textbf{pl}   & \textbf{gl}   & \textbf{uk}   \\ \midrule
JS w/ $\tau=1$ & \multicolumn{1}{c|}{\textbf{16.1}} & \textbf{13.8} & \textbf{7.2}  & \textbf{12.5} & \textbf{24.1} & \textbf{19.9} & \textbf{16.1} & \textbf{11.6} & \textbf{23.2} \\
JS w/ $\tau=3$ & \multicolumn{1}{c|}{16.3}          & 14.1          & 7.5           & 13.1          & 23.5          & 21.1          & 16.2          & \textbf{11.6} & 23.6          \\ \midrule
\multicolumn{10}{c}{\textbf{FLEURS (out-of-domain)}}                                                                                                                                 \\ \midrule
                & \multicolumn{1}{c|}{\textbf{avg}}  & \textbf{ca}   & \textbf{th}   & \textbf{ta}   & \textbf{hu}   & \textbf{cs}   & \textbf{pl}   & \textbf{gl}   & \textbf{uk}   \\ \midrule
JS w/ $\tau=1$ & \multicolumn{1}{c|}{\textbf{22.8}} & \textbf{15.4} & \textbf{15.1} & 21.6          & 37.2          & \textbf{29.8} & \textbf{21.4} & \textbf{16.7} & \textbf{25.1} \\
JS w/ $\tau=3$ & \multicolumn{1}{c|}{23.4}          & 17.0          & 15.6          & \textbf{21.5} & \textbf{36.0} & 31.4          & 22.4          & 16.8          & 26.2          \\ \bottomrule
\end{tabular}
\end{table}

Tables \ref{tab:results4} and \ref{tab:results5} present the results of comparing different temperatures (1 or 3) with the Jensen–Shannon loss for both the 10k and 28k settings. These results reveal that using a temperature of 1 generally results in improved in-domain and out-of-domain performance compared to a temperature of 3. However, for Tamil and Hungarian, temperature 3 showed better out-of-domain performance. These results suggest that \texttt{whisper-large-v2} serves as an effective teacher, justifying the use of a temperature of 1. Nevertheless, the optimal temperature value may vary depending on the quality of the teacher model for each specific language.

\begin{table}[caption={WER~($\downarrow$) for different training data sizes~(3k, 10k, and 28k utterances) for JS and KL losses for temperatures 1 and 3 for both in-domain~(CV-13) and out-of-domain~(FLEURS) test sets. Best results in \textbf{bold}.},label=tab:results5]
\begin{tabular}{@{}lcc|ccc|ccc@{}}
\toprule
                & \textbf{FLEURS} & \textbf{CV-13} & \multicolumn{3}{c|}{\textbf{FLEURS (out-of-domain)}} & \multicolumn{3}{c}{\textbf{CV-13 (in-domain)}} \\ \midrule
                & \textbf{avg}    & \textbf{avg}   & \textbf{ca}      & \textbf{ta}     & \textbf{th}     & \textbf{ca}    & \textbf{ta}    & \textbf{th}  \\ \midrule
JS w/ $\tau=1$ & \textbf{15.4}   & \textbf{9.3}   & \textbf{13.1}    & 19.2            & 14.0            & \textbf{11.3}  & \textbf{10.9}  & \textbf{5.7} \\
JS w/ $\tau=3$ & 16.3            & 9.7            & 14.8             & 20.1            & 14.1            & 11.8           & 11.3           & 5.9          \\
KL w/ $\tau=1$ & 15.6            & 10.8           & 14.6             & \textbf{18.7}   & \textbf{13.3}   & 14.9           & 11.3           & 6.2          \\
KL w/ $\tau=3$ & 16.5            & 9.7            & 15.8             & 19.8            & 14.0            & 12.2           & 11.1           & 5.9          \\ \bottomrule
\end{tabular}
\end{table}

Table \ref{tab:results5} also compares the use of the Jensen–Shannon (JS) loss with the traditional Kullback–Leibler (KL) loss discussed in Section \ref{sec:kd-background}, specifically for the 28k setting. Once again, the results favor a temperature of 1 in both cases, with a slight advantage for the JS loss against KL, primarily driven by Catalan out-of-domain performance. This advantage is more pronounced in in-domain performance. These findings indicate the presence of the \textit{mode-averaging problem} introduced in Section \ref{sec:kd-background}, although they are not definitive. They raise questions about whether these behaviors change when working with larger or smaller fine-tuning datasets and different levels of language resourcefulness. Unfortunately, due to time constraints, we could not explore these aspects in this study, leaving them as potential directions for future research.

\section{Multi-domain training}

In our final experiment, we delve into the impact of incorporating the train split of FLEURS dataset into our training data in the previously explored settings. The objective here is to use the validated architecture to generate models that would be more beneficial to the scientific community. In real-world scenarios, the models developed here are likely to be utilized in domains other than FLEURS or CV-13, so the hypothesis is that training on more than one dataset yields a better model.

\begin{table}[caption={WER~($\downarrow$) for the setting trained with 10k from CV-13 and FLEURS with dataset averages~(avg) for baselines~(top), adaptation approaches~(middle), and our method~(bottom) CV-13 and FLEURS test sets (both in-domain). Best results for \texttt{whisper-small} in \textbf{bold}.},label=tab:results6]
\resizebox{\textwidth}{!}{
\begin{tabular}{lcccccccccc}
\hline
\multicolumn{11}{c}{\textbf{Common voice 13.0}}                                                                                                                                                                              \\
\multicolumn{1}{c}{\textbf{Model}} & \textbf{Train data} & \multicolumn{1}{c|}{\textbf{avg}}  & \textbf{ca}   & \textbf{th}  & \textbf{ta}   & \textbf{hu}   & \textbf{cs}   & \textbf{pl}   & \textbf{gl}   & \textbf{uk}   \\ \hline
whisper large-v2                   & -                   & \multicolumn{1}{c|}{14.9}          & 16.9          & 9.3          & 17.3          & 18.6          & 14.5          & 8.1           & 19.0          & 15.6          \\
whisper-small                      & -                   & \multicolumn{1}{c|}{31.4}          & 30.1          & 20.3         & 30.1          & 45.5          & 38.6          & 18.8          & 35.7          & 32.3          \\ \hline
DistilWhisper                      & CV10k               & \multicolumn{1}{c|}{16.1}          & 13.8          & 7.2          & 12.5          & 24.1          & 19.9          & 16.1          & 11.6          & 23.2          \\ \hline
$\quad$+CLSR-FT                           & CV10k + F           & \multicolumn{1}{c|}{15.5}          & 15.1          & 6.8          & 12.4          & 21.9          & 18.4          & 16.3          & 11.3          & 22.2          \\
DistilWhisper                      & CV10k + F           & \multicolumn{1}{c|}{\textbf{14.6}} & \textbf{13.2} & \textbf{6.4} & \textbf{11.6} & \textbf{21.6} & \textbf{15.3} & \textbf{15.8} & \textbf{11.2} & \textbf{21.6} \\ \hline
\multicolumn{11}{c}{\textbf{FLEURS}}                                                                                                                                                                                         \\
\multicolumn{1}{c}{\textbf{Model}} & \textbf{Train data} & \multicolumn{1}{c|}{\textbf{avg}}  & \textbf{ca}   & \textbf{th}  & \textbf{ta}   & \textbf{hu}   & \textbf{cs}   & \textbf{pl}   & \textbf{gl}   & \textbf{uk}   \\ \hline
whisper large-v2                   & -                   & \multicolumn{1}{c|}{12.6}          & 5.6           & 12.6         & 19.3          & 17.9          & 14.4          & 5.9           & 16.8          & 8.3           \\
whisper-small                      & -                   & \multicolumn{1}{c|}{29.2}          & 14.6          & 22.7         & 36.2          & 42.9          & 40.3          & 18.2          & 33.5          & 24.8          \\ \hline
DistilWhisper                      & CV10k               & \multicolumn{1}{c|}{22.8}          & 15.4          & 15.1         & 21.6          & 37.2          & 29.8          & 21.4          & 16.7          & 25.1          \\ \hline
$\quad$+CLSR-FT                           & CV10k + F           & \multicolumn{1}{c|}{17.2}          & \textbf{11.8} & 10.1         & 16.0          & 28.1          & 23.2          & \textbf{17.1} & 12.9          & 18.7          \\
DistilWhisper                      & CV10k + F           & \multicolumn{1}{c|}{\textbf{16.7}} & 11.9          & \textbf{9.4} & \textbf{14.6} & \textbf{27.7} & \textbf{22.1} & 17.7          & \textbf{12.7} & \textbf{17.3} \\ \hline
\end{tabular}
}
\end{table}

Table \ref{tab:results6} showcases the outcomes of training the model in a setting involving 10k sentences from CV-13 along with the entire FLEURS train split. In this setting, we once again experiment with CLSR fine-tuning. For reference, the table also presents results from section \ref{sec:distilwhispervsothers}. The results reaffirm better performance for the setting with Knowledge Distillation compared to CLSR-FT. More significantly, the results demonstrate a substantial improvement within the domain when FLEURS is incorporated as part of the training dataset. Training with FLEURS reduces the WER on CV-13 by 1.5. This improvement is likely due to FLEURS' greater sentence complexity and larger average token count per line, contributing to enhanced training data diversity.

In table \ref{tab:results7}, we repeat the same experiment using settings with 3k and 28k sentences from CV-13, both added to the full FLEURS dataset. The results allow us to draw the same conclusions: the addition of out-of-domain training data (FLEURS) results in superior in-domain generalization on CV-13. Nevertheless, it is evident that the size of the training data remains a limiting factor, as CV3k+F (approximately 6k sentences) was insufficient to surpass CV10k alone, and similarly for CV10k+F (around 13k sentences) in comparison to CV28k alone.

In this section, we have presented the best models attainable for each setting using these two datasets. These models will be made open-source, and we hope they contribute to the development of speech recognition applications in these languages.

\begin{landscape}
\begin{table}[caption={WER~($\downarrow$) for the 3k setting with dataset averages~(avg) for baselines~(top), and our method~(bottom) for in-domain~(CV-13, FT only - higher portion) and out-of-domain~(FLEURS, all - middle portion) test sets. On the lower portion, the same results are grouped by resourcefulness. Best results for \texttt{whisper-small} in \textbf{bold}.},label=tab:results7]

\resizebox{\columnwidth}{!}{
\begin{tabular}{lccccccccccccccccc}
\hline
\multicolumn{1}{c}{}          &                     &                                    & \multicolumn{15}{c}{\textbf{Common voice 13.0}}                                                                                                                                                                                               \\
\multicolumn{1}{c}{\textbf{}} & \textbf{Train data} & \multicolumn{1}{c|}{\textbf{avg}}  & \textbf{bg}   & \textbf{ca}   & \textbf{cs}   & \textbf{fi}   & \textbf{gl}   & \textbf{hi}   & \textbf{hu}   & \textbf{id}   & \textbf{pl}   & \textbf{ro}   & \textbf{sk}   & \textbf{sl}   & \textbf{ta}   & \textbf{th}   & \textbf{uk}   \\ \hline
whisper large-v2              & -                   & \multicolumn{1}{c|}{17.0}          & 19.9          & 16.9          & 14.5          & 14.4          & 19.0          & 24.6          & 18.6          & 8.5           & 8.1           & 15.8          & 31.9          & 20.6          & 17.3          & 9.3           & 15.6          \\
whisper-small                 & -                   & \multicolumn{1}{c|}{34.2}          & 44.8          & 30.1          & 38.6          & 30.5          & 35.7          & 43.6          & 45.5          & 22.5          & 18.8          & 33.2          & 42.0          & 45.5          & 30.1          & 20.3          & 32.3          \\ \hline
DistilWhisper                 & CV3k                & \multicolumn{1}{c|}{22.6}          & 25.9          & 18.9          & 26.2          & 24.8          & 14.7          & 18.3          & 31.0          & 18.6          & 18.6          & 21.5          & 36.8          & 27.7          & 15.9          & 9.6           & 30.0          \\ \hline
DistilWhisper                 & CV3k + F            & \multicolumn{1}{c|}{\textbf{19.3}} & \textbf{21.8} & \textbf{15.0} & \textbf{21.7} & \textbf{22.4} & \textbf{14.2} & \textbf{15.8} & \textbf{26.4} & \textbf{17.0} & \textbf{17.2} & \textbf{18.0} & \textbf{29.3} & \textbf{22.9} & \textbf{13.4} & \textbf{7.8}  & \textbf{27.0} \\ \hline
\multicolumn{1}{c}{}          &                     &                                    & \multicolumn{15}{c}{\textbf{FLEURS}}                                                                                                                                                                                                          \\
\multicolumn{1}{c}{\textbf{}} & \textbf{Train data} & \multicolumn{1}{c|}{\textbf{avg}}  & \textbf{bg}   & \textbf{ca}   & \textbf{cs}   & \textbf{fi}   & \textbf{gl}   & \textbf{hi}   & \textbf{hu}   & \textbf{id}   & \textbf{pl}   & \textbf{ro}   & \textbf{sk}   & \textbf{sl}   & \textbf{ta}   & \textbf{th}   & \textbf{uk}   \\ \hline
whisper large-v2              & -                   & \multicolumn{1}{c|}{13.7}          & 14.6          & 5.6           & 14.4          & 9.7           & 16.8          & 23.8          & 17.9          & 7.1           & 5.9           & 14.4          & 11.7          & 23.1          & 19.3          & 12.6          & 8.3           \\
whisper-small                 & -                   & \multicolumn{1}{c|}{32.8}          & 39.9          & 14.6          & 40.3          & 26.8          & 33.5          & 47.9          & 42.9          & 18.6          & 18.2          & 34.6          & 35.8          & 54.5          & 36.2          & 22.7          & 24.8          \\ \hline
DistilWhisper                 & CV3k                & \multicolumn{1}{c|}{29.2}          & 42.8          & 17.2          & 35.6          & 32.0          & 18.7          & 21.8          & 41.1          & 19.1          & 21.9          & 33.1          & 35.2          & 50.5          & 25.7          & 17.6          & 25.9          \\ \hline
DistilWhisper                 & CV3k + F            & \multicolumn{1}{c|}{\textbf{18.3}} & \textbf{21.0} & \textbf{12.3} & \textbf{24.2} & \textbf{19.7} & \textbf{13.9} & \textbf{13.5} & \textbf{29.0} & \textbf{13.0} & \textbf{16.6} & \textbf{21.4} & \textbf{19.4} & \textbf{27.5} & \textbf{15.1} & \textbf{10.3} & \textbf{18.1} \\ \hline
\end{tabular}
}

\bigskip

\begin{tabular}{lccc|ccc|ccc}
\hline
\multicolumn{1}{c}{} & \textbf{Training} & \textbf{FLEURS} & \textbf{CV-13} & \multicolumn{3}{c|}{\textbf{FLEURS}}         & \multicolumn{3}{c}{\textbf{CV-13}}           \\
\multicolumn{1}{c}{} & \textbf{Data}     & \textbf{avg}    & \textbf{avg}   & \textbf{ca}   & \textbf{ta}   & \textbf{th}  & \textbf{ca}   & \textbf{ta}   & \textbf{th}  \\ \hline
whisper large-v2     & -                 & 12.5            & 14.5           & 5.6           & 19.3          & 12.6         & 16.9          & 17.3          & 9.3          \\
whisper-small        & -                 & 24.5            & 26.8           & 14.6          & 36.2          & 22.7         & 30.1          & 30.1          & 20.3         \\ \hline
DistilWhisper        & CV28k             & 15.4            & 9.3            & 13.1          & 19.2          & 14.0         & 11.3          & 10.9          & 5.7          \\ \hline
DistilWhisper        & CV28k + F         & \textbf{11.4}   & \textbf{9.0}   & \textbf{10.8} & \textbf{14.2} & \textbf{9.4} & \textbf{10.9} & \textbf{10.5} & \textbf{5.6} \\ \hline
\end{tabular}

\end{table}

\end{landscape}


            \chapter{Conclusion} \label{chap:conclusion}

This internship focused on investigating bias on Whisper, a family of large speech models, specifically examining speaker-related (gender, age, accent) and model-related (model size, resourcefulness, similar languages) biases. Additionally, we explored whether these biases are mitigated or exacerbated by quantization and proposed an alternative compression approach.

Our findings revealed that Whisper exhibits both speaker-related and model-related biases. Speaker-related biases are kept unchanged after quantization, while model-related biases are amplified by this compression technique. Low-resource languages are particularly more affected, and smaller models experience significant performance degradation. This is concerning because current parameter-efficient approaches typically apply quantization uniformly across models, introducing unintended bias.

To address this challenge, we introduced \textit{DistilWhisper}, a parameter-efficient distillation approach that enhances the performance of \texttt{whisper-small} by transferring the robustness of \texttt{whisper-large-v2} into a smaller model. This is achieved by incorporating language-specific gated modules and jointly optimizing ASR fine-tuning and knowledge distillation losses. Our results consistently showed performance improvements across various languages and test sets, with minimal parameter increase during inference. We believe this approach will democratize the use of Whisper models, making them accessible to a wider audience of researchers and practitioners. This approach was organized as a paper submitted to the conference ICASSP 2024~\citep{icassp2024-paper}. Code and models produced in this study will be made available soon on Hugging Face and Github. 

\section{Future Work}

There are several promising directions for future research in this area. Firstly, it would be beneficial to expand upon the analysis presented in Chapter \ref{chap:bias}, including an investigation into other quantization methods, such as 4-bit quantization. Exploring these methods across various model families would help determine if the conclusions drawn here are applicable more broadly. This could present an important contribution to the community and ensure the correct usage of these techniques.

Additionally, further research into the \textit{DistilWhisper} approach could yield valuable insights. Examining the effects of several hyperparameters, such as gate budget, KD loss weight, and temperature, would provide a deeper understanding of the approach's behavior. This exploration could help find the best setting for optimal performance of the approach.

Furthermore, it would be valuable to assess the impact of the proposed approach in multitasking beyond transcription (ASR), particularly in speech translation. Investigating whether language-specific paths can enhance translation performance to English, and exploring the potential for achieving new zero-shot capabilities in many-to-many translation scenarios, could open up exciting possibilities for the field.
    \end{content}
    
    \pagenumbering{Roman}
    \setcounter{page}{\numexpr\value{savepage}}

    \references{}
    
     \begin{appendix}


\section{Paper submitted to ICASSP 2024: "Multilingual DistilWhisper: Efficient Distillation of Multi-task Speech Models via Language-Specific Experts"} \label{app:paper-icassp}

Paper submitted to ICASSP 2024. 




     \end{appendix}

    
    
    
\end{document}